\documentclass[11pt]{article}
\usepackage[hypertexnames=false]{hyperref}
\usepackage{color}
\usepackage[dvipsnames]{xcolor}
\usepackage{latexsym}
\usepackage{dsfont}
\usepackage{appendix}
\usepackage{amssymb}
\usepackage{subcaption}
\usepackage[margin=.9in]{geometry}
\usepackage{float}
\usepackage{algorithm,algorithmic}
\usepackage[numbers]{natbib}
\usepackage{amsmath, amsfonts,amssymb,euscript,array,enumerate,amsfonts,mathrsfs}
\usepackage[dvipsnames]{xcolor}
\usepackage{amsthm}

\usepackage[T1]{fontenc}
\usepackage{hhline}
\usepackage{graphicx}
\usepackage{verbatim}
\usepackage{eurosym}
\usepackage{dsfont}
\usepackage{tabularx}

\allowdisplaybreaks

\makeatletter
\@addtoreset{equation}{section}

\newcommand*{\rom}[1]{\expandafter\@slowromancap\romannumeral #1@}
\makeatother
\newcommand{\vertiii}[1]{{\left\vert\kern-0.25ex\left\vert\kern-0.25ex\left\vert #1 
    \right\vert\kern-0.25ex\right\vert\kern-0.25ex\right\vert}}

\begin{document}
\title{On the Temperature of Machine Learning Systems}

\author{Dong Zhang
\thanks{\textbf{Email:} dongzhanghz@gmail.com }
}
\maketitle

\begin{abstract}

We develop a thermodynamic theory for machine learning (ML) systems.  Similar to physical thermodynamic systems which are characterized by energy and entropy, ML systems possess these characteristics as well. This comparison inspire us to integrate the concept of temperature into ML systems grounded in the fundamental principles of thermodynamics, and establish a basic thermodynamic framework for machine learning systems with non-Boltzmann distributions.  We introduce the concept of states within a ML system,  identify two typical types of state, and interpret model training and refresh as a process of state phase transition. We consider that the initial potential energy of a ML system is described by the model's loss functions, and the energy adheres to the principle of minimum potential energy. For a variety of energy forms and parameter initialization methods, we derive the temperature of systems during the phase transition both analytically and asymptotically, highlighting temperature as a vital indicator of system data distribution and ML training complexity.  Moreover, we perceive deep neural networks as complex heat engines with both global temperature and local temperatures in each layer. The concept of work efficiency is introduced within neural networks, which mainly depends on the neural activation functions. We then classify neural networks based on their work efficiency, and describe neural networks as two types of heat engines. 


\textbf{Keywords}: machine learning system, thermodynamics, temperature, entropy, energy, phase transition, heat engine, work efficiency

\end{abstract}


\section{Introduction}\label{section_intro}

From the perspective of information theory,  data carries entropy. The concept of entropy originated from thermodynamics and statistical mechanics, where it describes the disorder or randomness in a physical system. Claude Shannon later extended the idea of entropy to information theory to measure the uncertainty of random variables \cite{Shannon}, while Norbert Wiener also discussed entropy in the context of cybernetics, especially differential entropy \cite{cybernetics}. The employment of entropy in machine learning is an adaptation from information theory. For example, cross-entropy and information gain are used for splitting nodes in decision trees and random forests. In unsupervised learning, entropy can be used to evaluate the quality of clusters. Overall, the usage of entropy in data systems and machine learning is fundamentally rooted in the principles of thermodynamics and information theory, demonstrating a diverse and interdisciplinary application of the concept.

On the other hand, a physical system has energy, as well as entropy. If the concept of entropy can be introduced into a data system, does data also have energy? In the field of machine learning, there is a category of models known as energy-based models (EBMs) \cite{EBM_JMLR, EBL_lecun}. The origins of these EBMs can be traced back to the Ising model in statistical physics \cite{Ising, Ising_Intro} and the Amari-Hopfield network \cite{Amari,Hopfield}. The Boltzmann Machines (BMs) were proposed as stochastic recurrent neural networks \cite{BM}, inspired by the Ising model as well as spin-glass model in physics \cite{Spin_Glass}. To simplify the training process and improve computational efficiency, the Restricted Boltzmann Machines (RBMs) were later developed \cite{RBM01, RBM02}. The RBMs introduced a restriction that the neurons must form a bipartite graph, which significantly improved the training efficiency. Since the advent of RBMs, a variety of methods and applications have been proposed under the umbrella of EBMs, contributing to the evolution and expansion of this field \cite{Finn16, EBGAN, Mordatch18, RBM_QP, DuMordatch19, GEBM, train_EBM}. The fundamental concept of an EBM is to define an energy function that satisfies $E_{\mu} (x) = - \log p_{\mu}(x)$, or $p_{\mu} (x) = \exp [-E_{\theta}(x)/Z_{\theta}]$, where $E_{\mu}(x)$ is the energy function with parameter set $\mu$, and $Z_{\theta}$ is the partition function as the normalizing constant. This relationship between probability and energy aligns with the principles of statistical physics, and we can optimize either the loss function or the energy function to train the models. 



Of course, the so-called ``energy of data'' is not physical energy in the real world. Instead, it is an analogy drawn between data systems and the real physical world, serving as an indicator to describe the properties of data and machine learning systems associated with the data.  However, just like the introduction of entropy in information theory, we can consider the energy in data systems as a kind of generalized energy. Building on this concept, it leads us to consider: \textbf{if a machine learning (ML) system itself has energy, and incorporates the concept of entropy, the machine learning (ML) system can essentially be analogized to a thermodynamic system. } This raises an important question: \textbf{could we define temperature-like quantities to characterize the properties of a ML system?} The concept of temperature already exists in machine learning as a scaling parameter used to control the randomness of predictions made by models. For example, in BMs and RBMs, the temperature parameter appears in the Boltzmann distribution, with higher temperatures leading to more uniform distributions over states \cite{RBM_T1, RBM_T2, RBM_T3}. Similarly, the temperature parameter in softmax classifiers for multi-class classification is applied to the logits before the softmax function, with higher temperatures giving more similar probabilities to all classes \cite{softmax_T0, softmax_T1, softmax_T2}. Temperature can be used to control the creativity of a generative model, where higher temperatures will make more novel and unexpected predictions more likely \cite{GA_T1, GA_T2}. However, the existing concept of temperature in machine learning is merely a single model parameter, which cannot be derived from first principles, nor can it reflect the overall thermodynamic properties of the ML system. \textbf{ Overall, although the concepts of energy, entropy, and temperature exist in the field of machine learning, no one has yet unified the three concepts together, nor viewed ML systems as thermodynamic systems from first principles. } 

This paper systematically proposes a theory of thermodynamics and statistical mechanics in machine learning, with a particular focus on discussing the concept of temperature for ML systems. We can gain inspiration from the thermodynamic potentials in the real physical world. The thermodynamic potentials are fundamental concepts that describes the energy characteristics of a thermodynamic system. They are scalar quantities that provide information about the system state, and are used to understand how the system will respond to changes in temperature, pressure, and volume. There are several thermodynamic potentials, including internal energy ($U$), Helmholtz free energy ($F$), enthalpy ($H$) and Gibbs free energy ($G$). Correspondingly, we have a set of four equations known as the fundamental thermodynamic relations to describe these thermodynamic potentials, which are essential in understanding the behavior of thermodynamic systems \cite{Landau_stats}
\begin{eqnarray}
dU &= &TdS - PdV + \sum_i \mu_i dN_i \\
dF &= & -SdT - PdV + \sum_i \mu_i dN_i \\
dH & = & TdS + VdP + \sum_i \mu_i dN_i \\
dG & = & -SdT + VdP + \sum_i  \mu_i dN_i,
\end{eqnarray}
where $S$, $T$, $P$, $V$ are entropy, temperature, pressure and volume of the system respectively.  For fixed number of particles, volume or pressure, we have the following equations of state for temperature: 
\begin{equation}
T = \left(\frac{\partial U}{\partial S}\right)_{V, \{N_i\}} = \left(\frac{\partial H}{\partial S}\right)_{P, \{N_i\}}\label{eos_temp}.
\end{equation}

\begin{figure}[ht]
    \centering
    \includegraphics[width=1.0\linewidth]{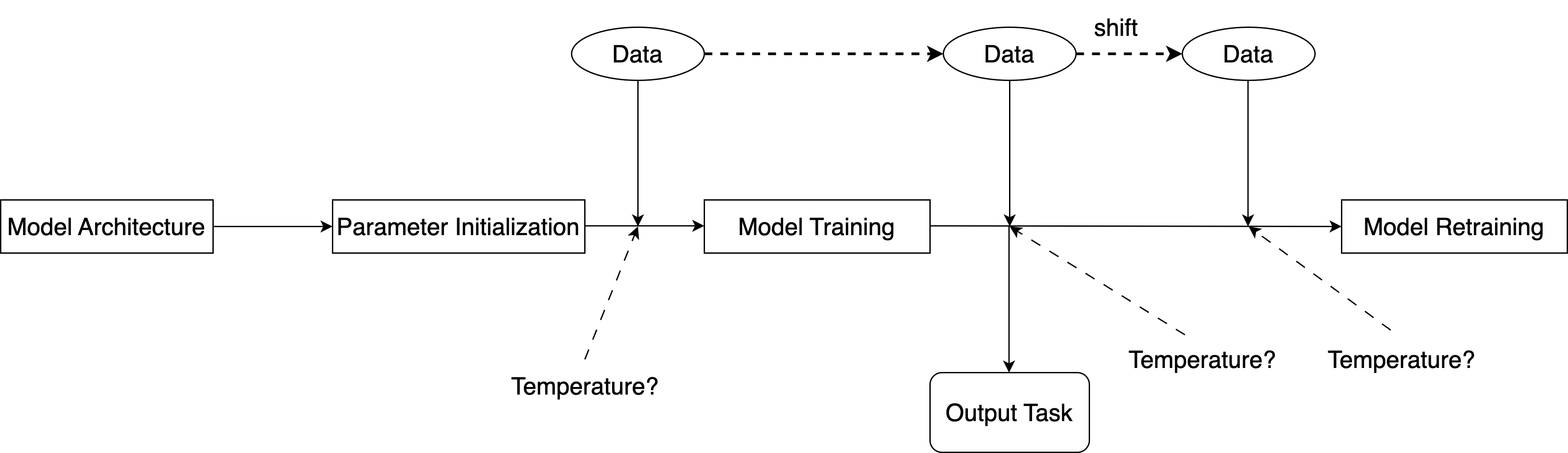}
    \caption{\small A machine learning (ML) system includes the initial model design, setting of initial parameters, importing data for training, importing new data for prediction tasks, and the process of keeping the model refreshed with new data. From a physics perspective, such a system with a series of steps is analogous to a heat engine. We can examine the various temperatures of the system during these processes, as well as the changes in energy and entropy.  }
    \label{fig_intro}
\end{figure}

On the other hand, a ML system has energy and entropy, but lacks some other physical quantities such as volume and pressure. We can use a method similar to equation (\ref{eos_temp}) to calculate the ML system temperature. Consider a system transitioning from one state to another, where the change in energy is $\Delta E$, and the change in entropy is $\Delta S$, the for an equilibrium system, we can use $T = \Delta E/\Delta S$ to calculate the temperature of the system. Note that for a machine learning system, we should not simply view it as a static data processor that goes from raw data input to prediction output. Instead, we should see it as a dynamic process, which includes a series of processes including the initial design of the model architecture, parameter initialization, optimization of the loss function and parameter tuning, and keep refreshing the model dynamically due to data shifts. Therefore, the temperature of a ML system must also be used to describe the entire process above. Figure \ref{fig_intro} shows such a system with multiple processes, and we can explore the various system temperature in different steps. 


This paper is organized as follows. In Section \ref{section_general_theory}, we develop the general theory to build a thermodynamic framework for ML systems. Three pivotal elements define an ML system: the data flow, model structure with its parameters, and system energy. In particular, we introduce two states of an ML system, corresponding to the stages of parameter initialization and data shifting respectively (Section \ref{section_state}). The ML training process can be viewed as an isothermal phase transition process, and the two states can be unified into a global picture (Appendix \ref{appendix_energy_1}). In Section \ref{section_system_energy}, we assign new physical meaning to the model loss function, viewing it as the internal potential energy of a ML system, which follows the principle of minimum potential energy. We emphasize that the meaning of energy in our theory differs from energy in energy-based models in literature. We then review system entropy in Section \ref{section_sys_entropy} and discuss the relation between discrete and differential entropy under dimension collapse scenarios (Appendix \ref{appendix_entropy_change}). The comparison of the ML system from an ML perspective versus a thermodynamic perspective is presented in Section \ref{section_comparison}.

Next, we derive the temperature of various ML systems based on different system energies and parameter initialization methods. In Section \ref{section_LR_MSE}, we develop the temperature theory in ML systems with a linear regression model and mean square error (MSE) as the internal energy, while parameters are initialized by normal distribution (Section \ref{section_MSE_normal}), uniform distribution (Section \ref{section_MSE_uniform}), and mixed distribution (Section \ref{section_MSE_mixed}). We also delve into alternative energy forms in ML systems, specifically the forms of Mean Square Error (MSE) and Mean Absolute Error (MAE) with regularization.  The temperatures for these systems are derived in  Sections \ref{section_MSE_regular} and \ref{section_MAE_linear} respectively.  Note that we prioritize deriving analytical expressions for system temperature. However, in cases where an analytical solution is elusive, we resort to asymptotic methods. We highlight the physical meaning of our temperature theory in Section \ref{section_physical_explan}, demonstrating temperature in equilibrium systems, and energy transfer and temperatures changes in non-equilibrium systems. Furthermore, we derive temperature in ML systems with logistic regression and cross-entropy energy in Section \ref{section_LR_CE}.

The thermodynamic theory of artificial neural networks (NNs) is developed in Section \ref{section_nn}. For simplicity, we only consider MSE energy and asymptotic solutions for NN systems in this paper. We discuss global and local temperature in NN systems, while the global temperature is for the whole system while local temperature is for each individual NN layer. In Section \ref{section_nn_heat}, we argue that an NN system can be viewed as a complex heat engine, where the engine work efficiency is defined as the ratio between the energy output from the last layer to the total energy released from the system. The system work efficiency is highly dependent on activation functions in NNs. Based on this, we define two types of heat engines categorized by their work efficiency.

\section{General Theory}\label{section_general_theory}

\begin{figure}[ht]
    \centering
    \includegraphics[width=0.4\linewidth]{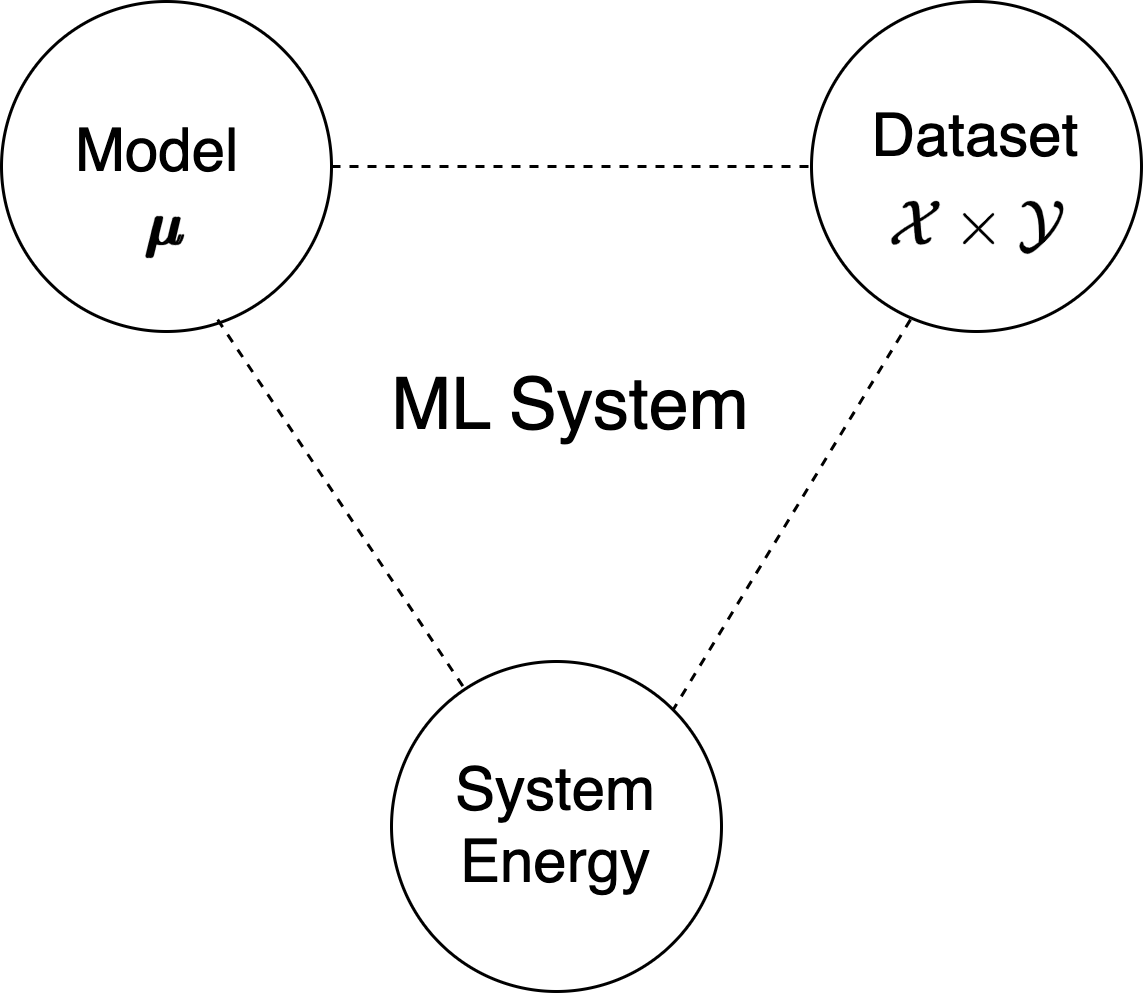}
    \caption{\small Three fundamental elements of a basic machine learning (ML) system.}
    \label{fig_ml_system}
\end{figure}

Let us first clarify the definition of ML system in this paper. Figure \ref{fig_ml_system} shows the three core elements of a basic ML system -- model, data, and energy. The model includes its architecture and parameters $\pmb \mu$. The data is used to train the model, optimize its parameters, and compare with the model output. The energy, as the intrinsic form of the system, will be studied from a thermodynamic perspective in this paper.

\subsection{State of a Machine Learning system}\label{section_state}

In thermodynamics, a thermodynamic state refers to the macroscopic state of a system that is characterized and specified by a set of measurable physical properties known as \textbf{state variables}. This definition of a thermodynamic state can be extended to ML systems. 

Let $\mathcal{X} \in \mathbb{R}^K$ denote a $K-$dimensional real valued random input vector, and $\mathcal{Y} \in \mathbb{R}$ a real valued random output variable, and the data domain of a ML system is $\mathcal {D} = \mathcal{X} \times \mathcal{Y}$.  A model function $f$ gives $\hat{ \mathcal{Y} } = f (\mathcal{X} )$, where $\hat{\mathcal{Y}}$ is the prediction which can be compared with the output variable $\mathcal{Y}$. From this we can define a class of functions
\begin{equation}
\mathfrak{F} = \{f | \hat{\mathcal{Y}}  = f_{\pmb \mu} (\mathcal{X}), {\pmb \mu}\in \mathbb{R}^n\}\label{equ_function_class},
\end{equation}
where ${\pmb \mu}$ is the set of parameters, which form the parameter space $\mathbb{R}^n$. A state in the ML system can be considered as a subset of $\mathcal X \times Y \times {\pmb \mu}$. 

In particular, we have two types of states as follows. 

\begin{figure}[ht]
    \centering
    \includegraphics[width=0.6\linewidth]{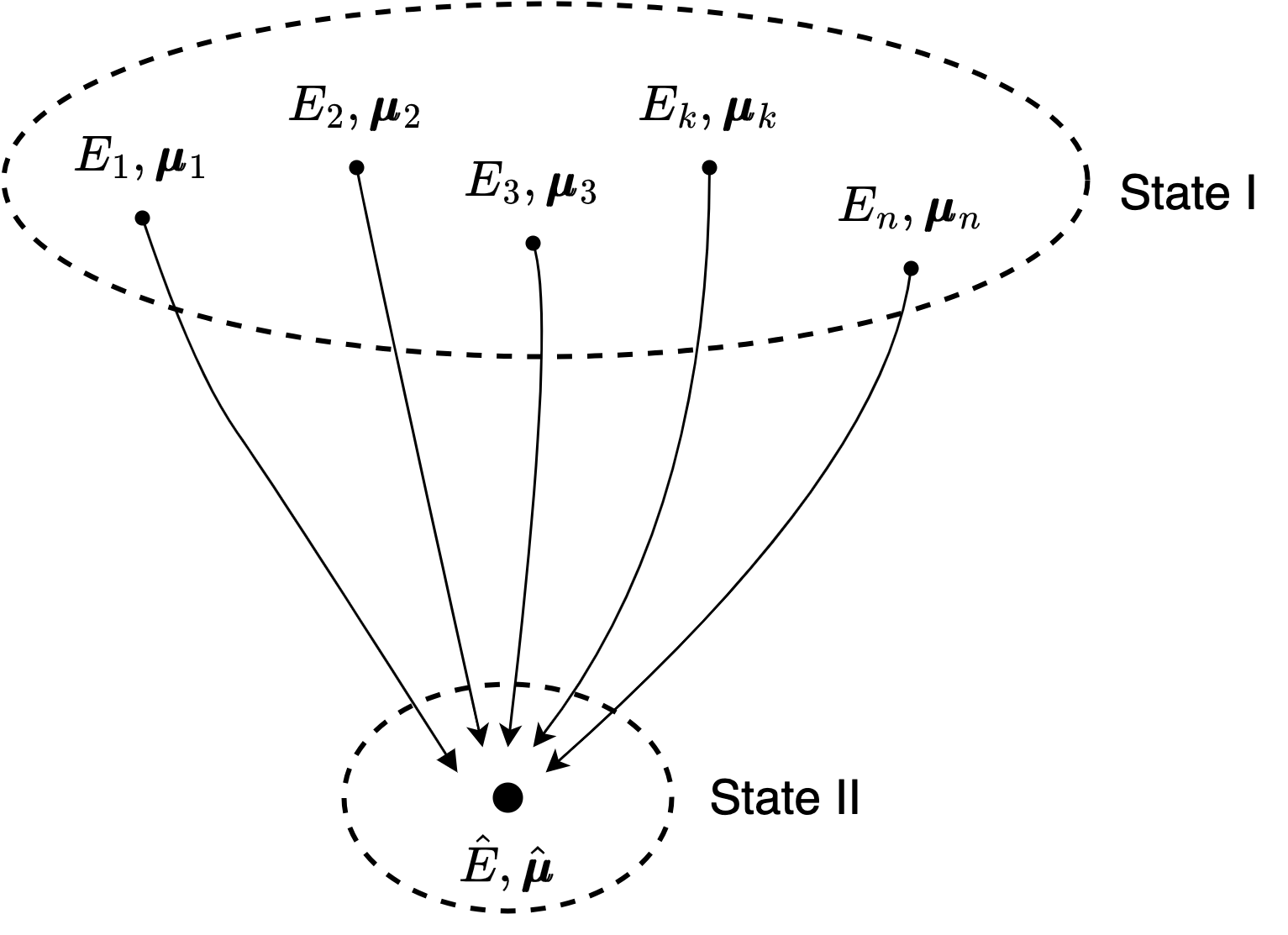}
    \caption{\small Type I state of a ML system (State I), which is the state for a given dataset with a set of parameters ${\pmb \mu}$. This state represents a ML system that has not yet been trained, and each $\{E_i, \pmb \mu_i\}$ can be considered as a particle. After training, all the particles converge to  $\{\hat{E}, \hat{\pmb \mu}\}$, which is State II. The transition from State I to State II is the process of the ML system going from initial state to trained state. Meanwhile, it can also be viewed as a phase transition process from State I to State II. We can use an isothermal phase transition process to calculate the temperature of the system. }
    \label{fig_state_1}
\end{figure}

\paragraph{Type I State and Phase Transition.}  For a given dataset from the domain $\mathcal {D} = \mathcal X \times Y$, all possible ${\pmb \mu}$ in the parameter space forms a state. State I from Figure \ref{fig_state_1} shows an example of this kind of state. For a given dataset, we have a series of parameters ${\pmb \mu} = \{\pmb \mu_i\}_{i=1,2,3, .., n}$, where each $\pmb \mu_i$ represents a point in the parameter space, corresponding to a function $f_i$ in $\mathfrak{F}$ from Equation (\ref{equ_function_class}). Also, the parameter set $\pmb \mu_i$ and function $f_i$ correspond to an energy level $E_i$, which is a function of $\pmb \mu_i$ and $\mathcal {D}$. All of the $\{E_i, \pmb \mu_i\}$ form a state, and each $\{E_i, \pmb \mu_i\}$  can be considered as a ``particle" in the system. 

The total energy of the state can be calculated by
\begin{equation}
E_0 = \sum_{i=1}^n E_i p_i \label{equ_E0},
\end{equation}
where $p_i$ is the probability of $\mu_i$ in the parameter space. On the other hand, the (Sharon) entropy of the state is
\begin{equation}
S_0 = -\sum_{i=1} p_i \log p_i \label{equ_S0},
\end{equation}
which is the entropy in the parameter space. Since the dataset in $\mathcal{D}$ is given and fixed, $S_0$ gives the overall entropy of the system. 

Now let us consider the training process of the ML system. For any given initial parameter ${ \pmb \mu_i}$, the system converges to an optimized state with energy $\hat{E}$ and parameter $\hat{\pmb \mu}$ after training. Ideally, regardless of the initial parameters, the system will eventually converge to the same $\{\hat{E}, \hat{\pmb \mu}\}$, and the corresponding function $\hat{f}$ is the optimal function among all functions in the function class $\mathfrak{F}$ to fit the relation between $\mathcal{X}$ and $\mathcal{Y}$. 


State I in Figure \ref{fig_state_1} corresponds to the parameter initialization process of the ML system. The ML training process is essentially transitioning from the parameter initialization state to the parameter convergence state. In Figure \ref{fig_state_1}, this is the \textbf{phase transition} process from State I to State II. During the phase transition, the system energy changes from $E_0$ (equation \ref{equ_E0}) in State I to $\hat{E}$ in State II, and the system entropy changes from $S_0$ (equation \ref{equ_S0}) to zero. Clearly, the phase transition is also the process of entropy reduction. Assuming the phase transition is an isothermal process, we can calculate the temperature of the system as
\begin{eqnarray}
T &=& \frac{E_{\rm stage\;II} - E_{\rm stage\; I}}{S_{\rm stage\;II} - S_{\rm stage\;I}} \nonumber \\
& = & \frac{\hat{E} - \sum_{i=1}^n E_i p_i }{\sum_{i=1} p_i \log p_i }\label{equ_T_discrete}\label{equ_state_I_T}.
\end{eqnarray}
Note that Equation (\ref{equ_T_discrete}) and Figure \ref{fig_state_1} are for the discrete case. For the continuous case, we have
\begin{eqnarray}
&&E_0 = \int E({\pmb \mu}) p({\pmb \mu})d{\pmb \mu},\\
&&S_0 = -\int p({\pmb \mu}) \log[p({\pmb \mu})]d{\pmb \mu},\label{equ_diff_entropy}\\
&& T = \frac{\int E_{\pmb \mu} d{\pmb \mu} -\hat{E}}{-\int p({\pmb \mu}) \log[p({\pmb \mu})]d{\pmb \mu}}\label{equ_T_1},
\end{eqnarray}
where $p({\pmb \mu})$ is the probability density function of ${\pmb \mu}$. However, it should be aware of that Equation (\ref{equ_diff_entropy}) gives differential entropy, which is fundamentally different from the discrete Shannon entropy. We will have more detailed discussion in Section \ref{section_sys_entropy}.

\begin{figure}[ht]
    \centering
    \includegraphics[width=0.8\linewidth]{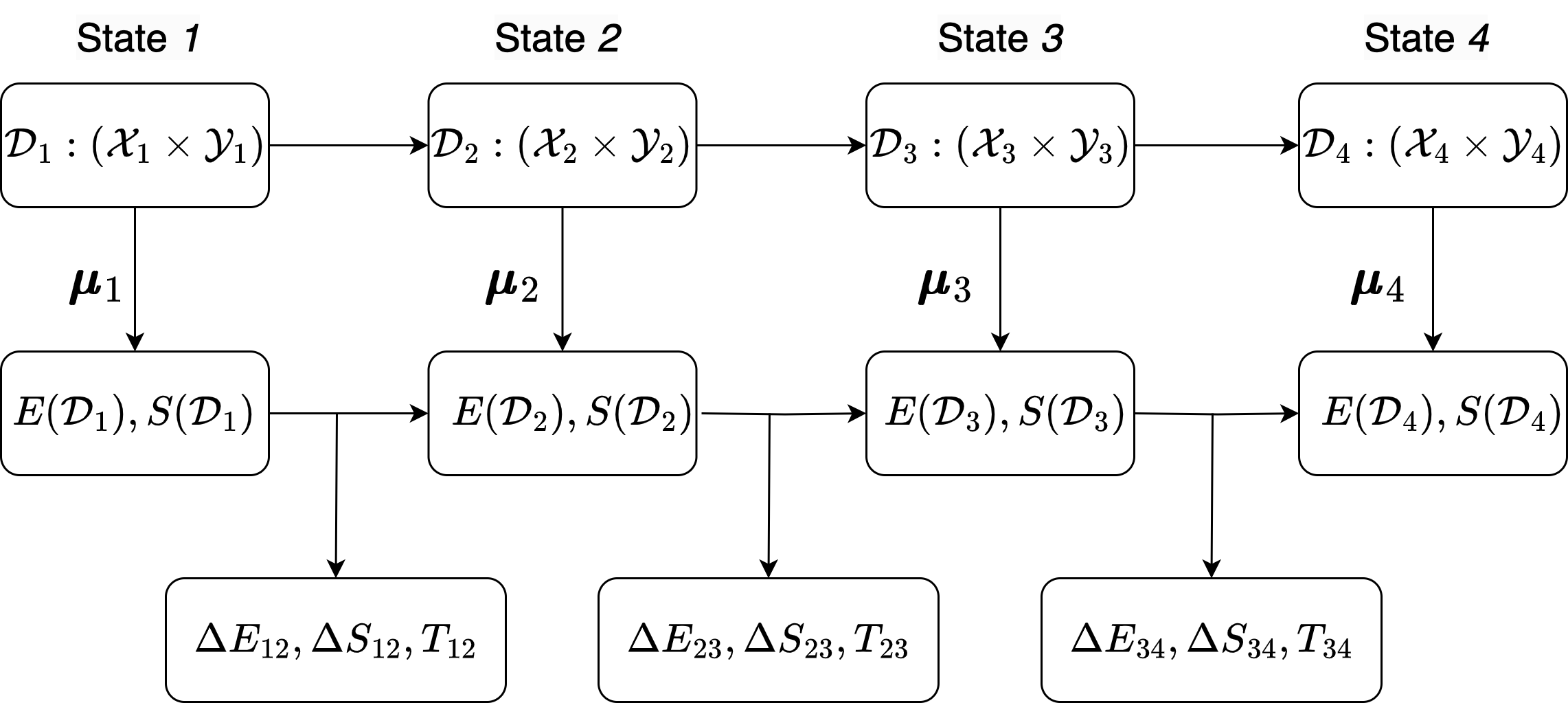}
    \caption{\small The evolution of the Type II state of a ML system (discrete case). The training dataset of the system is constantly evolving from $\mathcal{D}_1 = \mathcal{X}_1 \times \mathcal{Y}_1$ to $\mathcal{D}_2 = \mathcal{X}_2 \times \mathcal{Y}_2$ to $\mathcal{D}_3$..., forming a sequence ${\mathcal{D}_k}$, and the fixed parameter for the model ${\pmb \mu_k}$ and $\mathcal{D}_k$ give a Type II state with energy $\hat{E}(\mathcal{D}_k)$, and the entropy corresponding to the training dataset can be calculated using Equation (\ref{equ_stateII_entropy}). As the Type II state evolves, we can calculate the temperature of the state and its evolution through the changes in energy and entropy. }
    \label{fig_state_2}
     \vspace{-5pt}
\end{figure}

\paragraph{Type II State.} Different from the Type I state, the Type II state of a ML system is a state given by shifiting of the dataset after specifying a parameter set $\hat{\pmb  \mu}$. Assume the ML system's model is trained on a dataset $\mathcal{D}_1$, where $\mathcal{D}_1$  is a finite sequence of pairs in the data domain $\mathcal{X} \times \mathcal{Y}$ that $\mathcal{D}_1 = \mathcal{X}_1 \times \mathcal{Y}_1$ with $\mathcal{X}_1 \subset \mathcal{X}$, $\mathcal{Y}_1 \subset \mathcal{Y}$, and $\mathcal{D}_1 \subset \mathcal{D}$.  We use $E(\mathcal{D}_1)$ to denote the energy of the system under this state. Thus, the entropy of the state can be written as
\begin{equation}
S(\mathcal{D}_1) = -\sum_{(x_i, y_i) \in \mathcal{D}_1} P(x_i, y_i) \log[P(x_i, y_i)]\label{equ_stateII_entropy},
\end{equation}
where $P(x_i, y_i)$ is the joint probability in the domain, and each $(x_i, y_i)$ can be considered as a particle in Type II state.  Note that $\mathcal{D}_1 = \mathcal{X}_1 \times \mathcal{Y}_1$ is just a sample from the entire domain $\mathcal{D} = \mathcal{X} \times \mathcal{Y}$. Suppose the data used by the ML system is constantly shifting: $\mathcal{D}_1 = \mathcal{X}_1 \times \mathcal{Y}_1$ (State 1) $\rightarrow \mathcal{D}_2 = \mathcal{X}_2 \times \mathcal{Y}_2$ (State 2) $\rightarrow \mathcal{D}_3 = \mathcal{X}_3 \times \mathcal{Y}_3$ (State 3) ......., then the ML system can be viewed as a thermodynamic system that continuously absorbs or releases energy, as shown in Figure \ref{fig_state_2}. When the dataset of the system changes from $\mathcal{D}_j$  to $\mathcal{D}_{j+1}$, and the energy changes from $E(\mathcal{D}_{j})$ to  $E(\mathcal{D}_{j+1})$, the temperature ($T_{j,j+1}$) of the system during this process is
\begin{equation}
T_{j,j+1} = \frac{E(\mathcal{D}_{j+1}) - E(\mathcal{D}_j)}{\sum_{(x_i, y_i) \in \mathcal{D}_j} P(x_i, y_i) \log[P(x_i, y_i)] - \sum_{(x_i, y_i) \in \mathcal{D}_{j+1}} P(x_i, y_i) \log[P(x_i, y_i)]}\label{equ_T12},
\end{equation}
where in Equation (\ref{equ_T12}), the denominator shows the change in the system entropy, and the numerator gives the change in the system energy.
 
Figure \ref{fig_state_2} shows the discrete changes of the Type II state. If the training dataset shifting goes through a continuous changing process, i.e., the streamed training dataset can be written as $\mathcal{D}(\lambda) = \mathcal{X}(\lambda) \times \mathcal{Y}(\lambda)$, where $\lambda$ represents a continuous parameter that governs the streamed shifting. In this way, the differential entropy of the system is
\begin{equation}
S(\lambda) = - \int P[\mathcal{X}(\lambda), \mathcal{Y}(\lambda)] \log[P(\mathcal{X}(\lambda), \mathcal{Y}(\lambda))] d\mathcal{X} d\mathcal{Y}\label{equ_stateII_entropy_2},
\end{equation}
and we can obtain the temperature 
\begin{equation}
T(\lambda) = \frac{dE(\mathcal{X}, \mathcal{Y})/d\lambda}{-\int \{1 + \log[P(\mathcal{X}, \mathcal{Y})]\}
\left(\frac{dP}{d\mathcal{X}}\frac{d\mathcal{X}}{d\lambda} + 
\frac{dP}{d\mathcal{Y}}\frac{d\mathcal{Y}}{d\lambda}\right) d\mathcal{X} d\mathcal{Y}}.
\end{equation}

Different from the phase transition of Type I state, Type II state can be seen as gradually change or continuously change. The biggest question with Type II state is how to calculate the joint probability $P(\mathcal{X}, \mathcal{Y})$. In most realistic cases, the probability distribution of the data domain and of the training dataset sample are unknown. In order to effectively calculate the temperature change for Type II state, we must use a model to approximate and infer $P(\mathcal{X}, \mathcal{Y})$.  A commonly used method is to employ Bayes' theorem such that
\begin{equation}
P(\mathcal{X,Y}) = P(\mathcal{Y|X}) P(\mathcal{X})
\end{equation}
with $P(\mathcal{Y|X})$ obtained from modeling, and additional assumptions on $P(\mathcal{X})$.  Appendix \ref{appendix_I} provides details on how to calculate $P(\mathcal{X}, \mathcal{Y})$, and how to unify Type I and Type II states together within a single framework. 

The temperature in the Type II state is highly dependent on the changes in the system data, while the temperature in the Type I state depends more on the method of system parameter initialization and the form of energy expression, and can be derived from first principles. \textbf{This paper gives a comprehensive discussion on global temperature for ML systems in this section, with subsequent  discussions focusing mainly on the Type I state.}




\subsection{System Energy}\label{section_system_energy}

\begin{figure}[ht]
    \centering
    \includegraphics[width=0.8\linewidth]{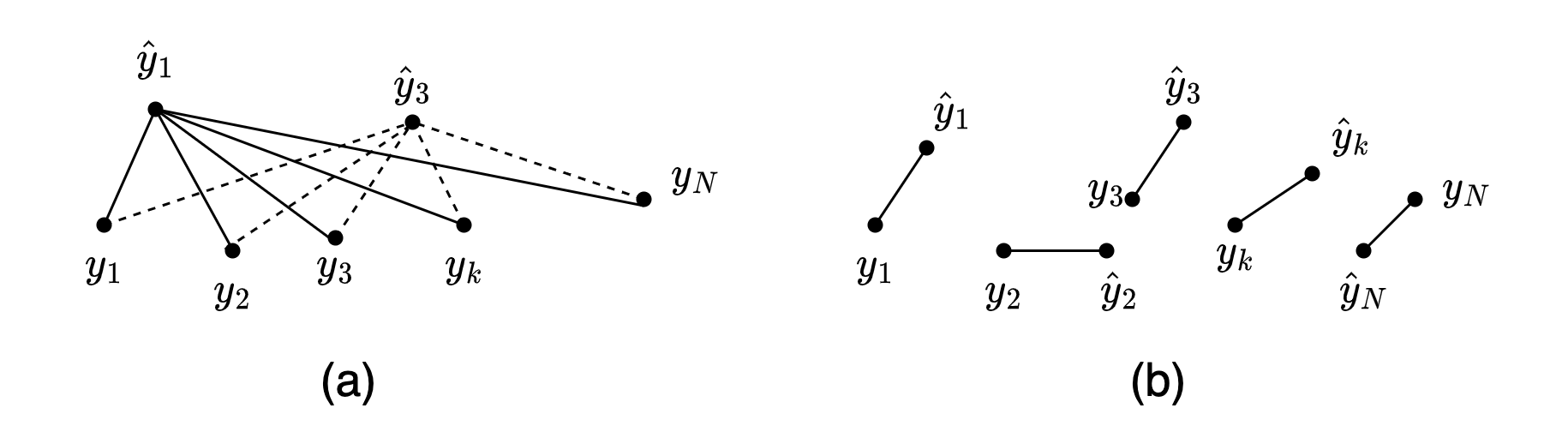}
    \caption{\small Energy of the system in $\mathcal{Y}$ space.  The left figure shows that the system has ``long-range energy", that is, there is energy $\mathscr{V}(y_i, y_j)$ between any internal particles $y_i$ and $y_j$ within the system. Also, there is energy $\mathscr{V}(\hat{y}_k,y_i)$ between any ``position" $\hat{y_k}$ and any internal particle $y_i$. See Equation (\ref{equ_energy_1}) to describe the energy of such as system. The right figure shows a simplified energy model where the system has ``short-range energy". In this model, any position within the system $\hat{y}_k$ interacts only with its nearest internal particle $y_k$. The mutual energy between internal particles is ignored due to its lack of variation. See Equation (\ref{equ_energy_2}) to describe such a system.}
    \label{fig_energy_1}
     \vspace{-5pt}
\end{figure}

In classical physics, the energy of a system can be divided into the potential and kinetic energy, the potential energy includes both the external background contributions, like gravitational potential energy, and the energy arising from particle interactions within the system. For example, for a molecular system, the potential energy of interaction between molecules can be described by the Lennard-Jones potential, which is written as \cite{LJ_potential_I, LJ_potential_II}
\begin{equation}
E_{L-J} = \sum_{i=1}^{N} \sum_{j>i}^{N} 4\epsilon \left[\left(\frac{\sigma}{r_{ij}}\right)^{12} -\left(\frac{\sigma}{r_{ij}}\right)^6 \right],
\end{equation}
where $r_{ij}$ is the distance between particles, and $\epsilon$  is the depth of the potential well. For a ML system, we can also construct a potential energy similar to the interaction between molecules. For systems with labels, the potential energy can be directly established in the $\mathcal{Y}$ space. Figure \ref{fig_energy_1}  shows demonstrations of such a potential. Let $\hat{y}_k = f_{\pmb \mu}(x_k)$, the left panel of Figure \ref{fig_energy_1} shows a general scenario for the potential energy of the system
\begin{equation}
E_p = \sum_{i=1}^N \sum_{k=1}^N \mathscr{V}(\hat{y}_k, y_i) + \sum_{i=1}^N \sum_{j>i}^N \mathscr{V}(y_j, y_i)\label{equ_energy_1},
\end{equation}
where the first term $\mathscr{V}(y_k, y_i)$ is the interaction energy between all positions $\hat{y}_k$ and $y_i$. As for the second term, if the system dataset remains unchanged, then this term is also constant, and thus we can ignore it. 

Equation (\ref{equ_energy_1}) accounts for the interactions among all particles in the system. However, if these interactions are short-ranged, a simplified approach is to only consider interactions between neighboring particles. This is often used in certain lattice gas models where particles on a lattice interact exclusively with their nearest neighbors. For example, in the Ising model, we consider a lattice where spins interact solely with their immediate neighbors. The right panel of Figure \ref{fig_energy_1} shows a similar concept within the context of a ML system, where $\hat{y}_i$ can be viewed as the displacement of $y_i$, and $y_i$ interacts only with $\hat{y}_i$. The interaction energy can be formulated as 
\begin{equation}
E_p = \sum_i \mathscr{V}(\hat y_i, y_i) = \sum_i ||f_{\pmb \mu}(y_i)  - y_i || \label{equ_energy_2}.
\end{equation}

Equation (\ref{equ_energy_2}) can be considered as the simplified version of Equation (\ref{equ_energy_1}), and is also a commonly used version of energy in machine learning. The most common forms of energy expressions are
\begin{itemize}
\item The mean squared error (MSE):
\begin{equation}
E_{\rm MSE} = \frac{1}{n}\sum _{i=1}^{n} (\hat{y}_i - y_i)^2 \label{equ_MSE},
\end{equation}
\item The mean absolute error (MAE):
\begin{equation}
E_{\rm MAE} = \frac{1}{n}\sum_{i=1}^{n}|\hat{y}_i - y_i|,
\end{equation}
\item The mean bias error (MBA):
\begin{equation}
E_{\rm MBA} = \frac{1}{n}\sum_{i=1}^{n}(\hat{y}_i - y_i),
\end{equation}
\item The Cross Entropy (CE):
\begin{equation}
E_{\rm CE} = -\frac{1}{n}\sum_{i=1}^{n} [y_i \log(\hat{y_i}) + (1-y_i) \log(1-\hat{y}_i)]\label{equ_energy_ce}.
\end{equation}
\end{itemize}

Different from EBMs, we tend to interpret the loss function as the potential energy of the system. In the context of EBMs, inference and learning refer to two distinct processes, where the goal of learning is to adjust the model parameters, while the goal of inference is to find configurations of the variables that optimize the energy function. For the state changes and phase transition described in Section \ref{section_state}, we actually view learning as the process of system energy change, so we also view the loss function corresponding to learning as the potential energy of the system. For example, loss functions with L1 or L2 regularization are
\begin{eqnarray}
{\rm L1:}\quad E_{p} = ||f_{\pmb \mu}(y_i) - y_i|| + \lambda \sum_j|\mu_j| \label{equ_L1}\\
{\rm L2:}\quad E_{p} = ||f_{\pmb \mu}(y_i) - y_i|| + \lambda \sum_j|\mu_j^2|. \label{equ_L2}
\end{eqnarray}
They can also be treated as two forms of potential energy. From a physical perspective, compared to Equation (\ref{equ_energy_2}), Equations (\ref{equ_L1}) and (\ref{equ_L2}) have extra terms added a ``background potential" that is unrelated to the internal potential of the system.

Another example is the energy for binary classification problem. In addition to the cross entropy energy in the aforementioned Equation (\ref{equ_energy_ce}), the energy for binary classification can also be expressed as (see Lecun et al. \cite{EBL_lecun}) $E= \sum_i |2 y_i - 1| f_{\pmb \mu}^{(i)}(x)$, where $f_{\pmb \mu}(x)$ is typically a linear combination of ${\pmb \mu}$ and $x$, i.e., ${\pmb \mu}x +{\pmb b}$. Clearly, we prefer to use the cross entropy expression as the energy model for binary classification.

From the perspective of the loss function, the setting of model parameters is determined by minimizing a chosen loss function, i.e., 
\begin{equation}
\hat {\pmb \mu} = {\rm argmin}_{\pmb \mu}\frac{1}{n}\sum_{i=0}^n \mathcal{L}(\hat{y}(x_i, {\pmb \mu}), y_i).
\end{equation}
This gives us a hint that training a model involves adjusting its parameters to minimize this loss function, much like finding the position where potential energy is minimized. \textbf{If we consider the loss function $\mathcal{L}$ as the internal potential energy of the system, we can view the process of optimizing the loss function through the lens of the principle of minimum potential energy}, which states that a stable physical system seeks to establish an equilibrium state that minimizes its potential energy $E_p$ such that when an infinitesimal variation from such position involves no change in energy $\delta E_p = 0$.

That is to say, if we consider the loss function as the system internal energy, then the process of optimizing the loss function, from a physics perspective, is equivalent to the process of the system minimizing its internal energy. This also shows the reasonableness of viewing the ML system as a physical system that conforms to physical laws.

It is important to clarify that the system energy we propose is fundamentally different from the energy in EBMs. Within the EBM framework, both the loss function and the energy function coexist. The energy function is minimized by the inference process, while the loss functional is minimized by the learning process. Differently, in our Type I state phase transition, as energy is released, the system parameters gradually change, while the data remains constant (see Figures \ref{fig_intro} and \ref{fig_ml_system}). This process corresponds to the traditional learning process. In our theoretical framework, we do not distinguish between the loss function and the energy function.

\subsection{System Entropy}\label{section_sys_entropy}


\paragraph{Discrete Entropy and Differential Entropy.} In this section, let us take a deep dive into the entropy changes of the Type I state system. Given a distribution in either parameter or data space, how can we determine the entropy of all particles? One direct approach is to segment the space into grids, and assess the probability of particles within each grid. Since the distribution for parameter initialization in parameter space and data spatial distribution derived from models are generally continuous, we can calculate the differential entropy of the system. However, given that the particle count in a ML system is always finite, it appears we should evaluate the particle entropy using discrete entropy. In this section, let us investigate the connection between differential entropy and discrete entropy, and confirm how to calculate entropy change in a ML system. 

Assume there are $N$ particles in the parameter space $\{\pmb \mu\}$, and the particles are independent of each other. Therefore, the entropy of the particles is the linear superposition of the entropy contributed by each particle. To calculate the contribution of a single particle, assume the parameter space is divided into a series of grids with the volume of the $i-$th grid being $\Delta_i$, and the probability of the particle in that grid is $p_i$, so the discrete entropy of the particle is
\begin{equation}
S_{\rm discrete} = -\sum_i p_i \log p_i \label{equ_entropy}.
\end{equation}
Assume the probability distribution of the particle is $f({\pmb \mu})$, we have $p_i = f({\pmb \mu}) \Delta_i$, and the above equation becomes $S_{\rm discrete} = -\sum_i f({\pmb \mu_i}) \Delta_i \log [f({\pmb \mu_i})\Delta_i]$, and for all $\Delta_i \rightarrow \delta \rightarrow 0$, We have obtained the connection between discrete and differential entropy as
\begin{eqnarray}
S_{\rm discrete} &=& -\int f({\pmb \mu}) \log [f({\pmb \mu})]d{\pmb \mu} - \log \delta \\
&=& S_{\rm diff} - \log \delta \label{equ_S_discrete},
\end{eqnarray}
where $S_{\rm diff}$ is differential entropy for the distribution $f({\pmb \mu})$. Thus, the entropy for $N$ particles is 
\begin{equation}
S_0 = N S_{\rm diff} - N \log \delta.
\end{equation}
On the other hand, the energy possessed by $N$ particles (before phase transition) is
\begin{eqnarray}
E_{\rm initial} &=& \sum_i E({\pmb \mu}_i) n_i \Delta_i = \sum_i E({\pmb \mu})(N f_{\pmb \mu}\Delta) \nonumber \\
&\rightarrow& N \int E({\pmb \mu})f({\pmb \mu})d{\pmb \mu}.
\end{eqnarray}
After the phase transition, the total energy is assumed to be $N E_f$ and the final discrete entropy drops to zero, so we have the temperature
\begin{equation}
T = \frac{N \int E({\pmb \mu})f({\pmb \mu})d{\pmb \mu} - N E_f}{N S_{\rm diff} - N \log \delta} = \frac{\int E({\pmb \mu})f({\pmb \mu})d{\pmb \mu} - E_f}{S_{\rm diff} - \log \delta},
\end{equation}
The term $\log\delta \rightarrow -\infty$ for $\delta \rightarrow 0$ is trivial. In Appendix \ref{appendix_entropy_change}, we discuss from a higher perspective that $\log \delta$ represents the dimensional collapse of the system, and $S_{\rm diff}$ shows the real entropy change for Type I state phase transition. Therefore, we can use the different entropy as the indicator of entropy change, and write
\begin{equation}
T = \frac{ E_f - \int E({\pmb \mu})f({\pmb \mu})d{\pmb \mu}}{\int f({\pmb \mu}) \log [f({\pmb \mu})]d{\pmb \mu}}\label{equ_T_diff},
\end{equation}
as shown in Equation (\ref{equ_T_1}) from Section \ref{section_state}.

\paragraph{Parameter Entropy and Data Entropy.} Let us distinguish between the concepts of parameter entropy and data entropy. The parameter entropy, denoted as $S_{\rm parameter}$, arises from the randomness of parameter initialization, while the data entropy $S_{\rm data}$, originates from the randomness of the system data. For a complete ML system, the system entropy $S_{\rm sys}$ has

\begin{equation}
S_{\rm sys} = S_{\rm parameter} + S_{\rm data}.
\end{equation}

For the Type I state, given that the data remains constant with a fixed $S_{\rm data}$, our primary focus is on the entropy contributed by parameter initialization, as described by Equation  (\ref{equ_S_discrete}).  Conversely, for the Type II state, since the parameters are already optimized with a fixed $S_{\rm parameter}$, but $S_{\rm data}$ varies due to data shifts, our attention is solely on data entropy, which can be obtained by  Equations (\ref{equ_stateII_entropy}) and (\ref{equ_stateII_entropy_2}).


\subsection{Non-Boltzmann Distribution}\label{section_non_boltzmann}

In much of the literature about energy-based models (EBMs), energy and probability distribution are linked together. In classical statistical, there is a relation between probability $P(x)$  and energy $E_{\pmb \mu}$, where  $P(x)$ is proportional to $\exp[-\beta E_{\pmb \mu}(x)]$ with $\beta = 1/T$, and $T$ being the temperature of the system. After normalization, we can get
\begin{equation}
P(x) = \frac{\exp[-E_{\pmb \mu}(x)/T]}{\int_{x\in\mathcal{X}} \exp[-E_{\pmb \mu}(x)/T]dx}\label{equ_boltzmann_1}
\end{equation}
or the conditional probability distribution
\begin{equation}
P(y|x) = \frac{\exp[-E_{\pmb \mu}(x,y)/T]}{\int_{y\in\mathcal{Y}}\exp[-E_{\pmb \mu}(x,y')/T]dy'}\label{equ_boltzmann_2}.
\end{equation}

Equations (\ref{equ_boltzmann_1}) and (\ref{equ_boltzmann_2}) have also been used in EBMs with $T$ as a scalar parameter for the models.  However, we must point out that equations (\ref{equ_boltzmann_1}) and (\ref{equ_boltzmann_2}) do not always hold true, even within the realm of statistical mechanics. In Appendix \ref{appendix_non_boltzmann}, we discusses the cases of \textbf{non-Boltzmann distributions}, which can also be referred to as \textbf{generalized Boltzmann distributions}. The probability distribution also depends on the non-thermodynamic properties of the system, such as gravity and the choice of reference frame. For most thermodynamic systems, the generalized Boltzmann distribution is written as
\begin{equation}
P(x) = \frac{f(\mathcal{G}_i) \exp[-E_{\pmb \mu}(x)/T]}{\int_{x\in\mathcal{X}}f(\mathcal{G}_i) \exp[-E_{\pmb \mu}(x)/T]dx}\label{equ_boltzmann_3},
\end{equation}
where $f(\mathcal{G})$  is a gravity-like factor to shape the  probability.


Hence, initiating from the classical Boltzmann distribution and trying to forge a link between energy through probability equations  (\ref{equ_boltzmann_1}) and (\ref{equ_boltzmann_2}) may not be the most effective approach. We would prefer to calculate the temperature of the system from first principles, specifically by assessing the changes in the system's  energy and entropy. In Appendix \ref{appendix_lorentz}, we briefly review research from the literature on the Lorentz transformation of temperature. In general, the fundamental thermodynamic equations are covariant in any reference frame, and the relation $T = \partial U/\partial S$ (equation \ref{eos_temp}) always holds. The principle is more fundamental than the Boltzmann distribution.

\subsection{ML vs Physics Perspective, and the Following Content}\label{section_comparison}

Based on the discussions above in Section \ref{section_general_theory}, we establish an analogy between thermodynamic systems and ML systems. Table \ref{tab_comparison} provides the terminology descriptions of various concepts in ML from a physics perspective. Throughout this paper, we will frequently use various physics terms. In this paper, we regard the terms from both the ML and physics perspectives as interchangeable and mutually equivalent. 

\begin{table}[ht]
\footnotesize
\centering
\begin{tabular}{lll}
\hline
\textbf{ML Perspective} & \textbf{Physics Perspective} & \textbf{Section(s)} \\
\hline
ML system & Thermodynamic system & Section \ref{section_intro} \\
\hline
ML system temperature & Physical temperature & Section \ref{section_intro} \\
\hline
Information entropy & Physical entropy & Section \ref{section_intro}, Section \ref{section_sys_entropy} \\
\hline
Differential vs. discrete entropy & Dimension collapse &  Section \ref{section_sys_entropy}, Appendix  \ref{appendix_entropy_change}\\
\hline
Model parameter initialization & Type I state particles& Section \ref{section_state}\\
\hline
Model training & Type I state phase transition & Section \ref{section_state} \\
\hline
System data shift & Type II state evolution& Section \ref{section_state} \\
\hline
Loss function & Internal potential energy & Section \ref{section_system_energy} \\
\hline
Loss functions optimization & Minimum potential energy & Section \ref{section_system_energy} \\
\hline
Parameter entropy & Type I state entropy & Section \ref{section_sys_entropy} \\
\hline
Data entropy & Type II state entropy & Section \ref{section_sys_entropy} \\
\hline
Model retraining & Thermodynamic equilibrium & Section \ref{section_physical_explan} \\
\hline
Entire process of ML (re)training cycles & Unified Scenario of Type I \& II states & Appendix \ref{appendix_I} \\
\hline
Neural network & Complex heat engine & Section \ref{section_nn_heat} \\
\hline
Tanh and Sigmoid activation & Low work efficiency (Type I heat engine) & Section \ref{section_nn_heat} \\
\hline
ReLU activation & High work efficiency (Type II heat engine) & Section \ref{section_nn_heat} \\
\hline
\end{tabular}
\caption{A comparison of various terms from the ML perspective and the physics perspective.  The last three pairs of terms are from the following Sections. The ML terms and physics terms are used interchangeable in this paper.}
\label{tab_comparison}
\end{table}

Next, let us calculate the temperature for a series of common machine learning systems. Based on the systems, they can be categorized as follows:

\begin{itemize}
    \item  \textbf{Parameter Initialization:}
The most common ways to initialize (model) parameters of the system are through normal distribution and uniform distribution.

   \item  \textbf{Energy Forms:}
Common energy forms, as mentioned in Section \ref{section_system_energy}, are borrowed from ML loss functions, which include Linear Regression (MSE, MAE, etc.), Logistic Regression (Cross Entropy), and original energy forms adjusted by $L_1$ and $L_2$ regularization.
\end{itemize}

\begin{table}[h]
\centering
\footnotesize
\begin{tabular}{llll}
\hline
\textbf{Energy Form} & \textbf{Initial Parameter Distribution} & \textbf{Method}& \textbf{Section} \\
\hline
\hline
Linear Regression with MSE & normal distribution & analytic& Section \ref{section_MSE_normal}\\
\hline
Linear Regression with MSE & uniform distribution & analytic& Section \ref{section_MSE_uniform}\\
\hline
Linear Regression with MSE & normal and uniform mixed & analytic& Section \ref{section_MSE_mixed}\\
\hline
Linear Regression with MSE and regularization & normal/uniform distribution& analytic& Section \ref{section_MSE_regular}\\
\hline
Linear Regression with MAE & normal/uniform distribution& asymptotic & Section \ref{section_MAE_linear}\\
\hline
Logistic Regression with Cross Entropy & normal distribution & asymptotic & Section \ref{section_CE_normal}\\
\hline
Logistic Regression with Cross Entropy & uniform distribution & asymptotic & Section \ref{section_CE_uniform}\\
\hline
Neural Network with MSE (entire system)& normal distribution & asymptotic & Section \ref{section_nn_normal}\\
\hline
Neural Network with MSE (each layer)& normal distribution & asymptotic & Section \ref{section_nn_layer}\\
\hline
Neural Network with MSE& uniform distribution & asymptotic & Section \ref{section_nn_others}\\
\hline
\end{tabular}
\caption{\small The temperatures of various ML systems. The energy forms of the systems include linear regression with MSE and MAE, logistic regression with Cross Entropy, and neural networks with MSE. The initial parameter distributions for the systems are either normal or uniform. We employed either analytical method or asymptotic approximations to investigate the ML systems and their temperatures.}\label{tab_structure}
\end{table}

Henceforce, the remainder of this paper will discuss the temperature of various ML systems based on the two aforementioned types of parameter initialization and various energy forms. In particular, we focus on the system temperature during the phase transition process of system State I. Table \ref{tab_structure} gives the various systems we will study.  Note that if the study ``Method'' in Table \ref{tab_structure} is ``analytic'', it means that the we can derive the system temperature analytically, while ``asymptotic'' means that the temperature does not have an analytical solution, and we consider the extreme cases, such as the standard deviation $\rightarrow \infty$ for the normal distribution, or the range of the uniform distribution expands to infinity in order to obtain an approximate solution for the  temperature.  The approximated asymptotic solutions still have clear meanings for the ML systems. In this paper we only consider analytical and asymptotic solutions, and do not expand into numerical solutions and simulations.

\section{Linear Regression with MSE}\label{section_LR_MSE}

\subsection{Parameter Initialization: Normal Distribution}\label{section_MSE_normal}

Assuming a ML system has an initial Type I state with set of parameters $\{\pmb \mu\}$ with $K$ dimensions, and the parameters are initialized by normal distribution. The differential entropy of  the multivariate Gaussian is
\begin{equation}
S = \frac{1}{2} \ln |\Sigma| + \frac{K}{2} [1+ \ln(2\pi)] \label{eq_normal_entropy},
\end{equation}
In our case we assume each dimension has independent distribution, i.e.,
\[
\Sigma = \left[
\begin{array}{ccccc}
\sigma_{1}^2 &  &  &  &\\
 & \sigma_{2}^2 & &  &\\
 &  & \sigma_{3}^2 &  &\\
  &  & & ... & \\
   &  & &   &  \sigma_{K}^2 
\end{array} 
\right],
\]
From Equation (\ref{eq_normal_entropy}) the initial entropy of the system becomes
\begin{equation}
S_0 = \ln (\sigma_1\sigma_2 ... \sigma_K) + \frac{K}{2} [1 + \ln (2\pi) ].
\end{equation}

\subsubsection{2D Linear Regression}\label{section_MSE_normal_2D}

Let us start from 2-dimensional linear regression with the loss function MSE  $\frac{1}{n} \sum(\mu_1 x_i + \mu_2 - y_i)^2$, where $n$ is the number of data points $\{x_i, y_i\}$  (see Equation [\ref{equ_MSE}]). As we initialize the parameter set with normal distribution, the average energy of the state is
\begin{eqnarray}
\langle E_0 \rangle &= & \int \frac{1}{n} \sum_i^{n}(\mu_1 x_i + \mu_2 -y_i)^2 \frac{1}{\sqrt{2 \pi}\sigma_1} \textrm{e}^{-\frac{\mu_1^2}{2 \sigma_1^2}} \frac{1}{\sqrt{2 \pi}\sigma_2} \textrm{e}^{-\frac{\mu_2^2}{2 \sigma_2^2}} d\mu_1 d\mu_2 \nonumber\\
&= & \int \frac{\sigma_2^2}{2\pi n} d\mu_1 \textrm{e}^{-\frac{\mu_1^2}{2}} \sum \int \left[\mu_2 + \left(\mu_1\frac{\sigma_1}{\sigma_2}x_i - \frac{y_i}{\sigma_2}\right) \right]^2 \textrm{e}^{-\frac{\mu_2^2}{2}} d\mu_2 \nonumber \\
& = & \sum \int \frac{\sigma_2^2}{\sqrt{2\pi} n} \textrm{e}^{-\frac{\mu_1^2}{2}} \left[\left(\mu_1 \frac{\sigma_1}{\sigma_2}x_i -\frac{y_i}{\sigma_2}\right)^2 + 1\right]d\mu_1 \nonumber \\
& = &  \sum \frac{\sigma_1^2 x_i^2}{n}\left[1 + \frac{y_i^2}{x_i^2 \sigma_1^2} + \frac{\sigma_2^2}{\sigma_1^2 x_i^2}\right] \nonumber \\
& = & \sum \frac{\sigma_1^2 x_i^2 + y_i^2 + \sigma_2^2}{n} \nonumber \\
& = & \sigma_1^2 \overline{X^2} + \overline{Y^2} + \sigma_2^2,
\end{eqnarray}
where $\overline{X^2} = \sum x_i^2/n$ and $\overline{Y^2} = \sum y_i^2/n$ are denoted as the mean of the squared values of the data in the system.  

After Type I state phase transition, the internal energy of the system goes to the minimum point that 
\begin{eqnarray}
E_f & = & \textrm{min}\left[\frac{1}{n}\sum(\mu_1 x_i + \mu_2 - y_i)^2\right] \nonumber \\
& = & \frac{\sum (y_i - \bar{y})^2}{n} - \frac{\sum[(x_i - \bar{x})(y_i - \bar{y})]^2}{n(x_i - \bar{x})^2} \nonumber \\
& = &  \textrm{Var}(Y) - \frac{\textrm{Cov}(X,Y)^2}{\textrm{Var}(X)} \nonumber \\
& = & (1 - \rho^2) \textrm{Var}(Y) \label{equ_MSE_Ef}.
\end{eqnarray}
Thus, the temperature of the system is
\begin{eqnarray}
T &=& \frac{\langle E_0 \rangle - E_f}{S_0} \nonumber \\ 
&=& \frac{ \sigma_1^2 \overline{X^2} + \overline{Y^2} + \sigma_2^2 - (1 - \rho^2) \textrm{Var}(Y)}{1 + \ln (2\pi \sigma_1 \sigma_2)}\label{equ_T_linear_MSE}
\end{eqnarray}
Equation (\ref{equ_T_linear_MSE}) gives the analytical solution of the system temperature for a linear regression MSE energy form with initial normal distribution. The temperature is a function of data distribution $\overline{X^2}$, $\overline{Y^2}$, as well as initial parameter distribution $\sigma_1$ and $\sigma_2$. 

For the asymptotic case  $\sigma_1, \sigma_2  \rightarrow \sigma \rightarrow \infty$,  where the parameter initialization distribution has the maximum randomness, we obtain the asymptotic property of temperature that
\begin{equation}
T \sim \frac{ \sigma^2 + \sigma^2 \overline{X^2} + \overline{Y^2}}{ 2\ln \sigma} \sim \left(\frac{\sigma^2}{2\ln  \sigma}\right) (1+ \overline{X^2})\label{equ_T_MSE_normal}.
\end{equation}

\subsubsection{High-dimensional Linear Regression}\label{section_MSE_normal_highD}

The 2D cases in Section \ref{section_MSE_normal_2D}  can be extended to higher dimensions. The energy form in high dimensions is written as
\begin{equation}
{\rm MSE} = \frac{1}{n} \sum_i \left(\sum_{j=1}^{K-1} \mu_j x_{ij} + \mu_{K} - y_i\right)^2,
\end{equation}
where $K$ is the dimension of the parameter space, and $x_{ij}$ is the $j$-th component of the $i$-th data point $\pmb x_i$,  Based on this, we can obtain the initial energy of the system
\begin{eqnarray}
\langle E_0 \rangle &= & \int \frac{1}{n}\sum_i \left(\sum_{j=1}^{K-1} \mu_j x_{ij} + \mu_{K} - y_i\right)^2 \frac{1}{(2 \pi )^{K/2}\prod _{j=1}^{K} \sigma_j}\textrm{e}^{-\sum_{j=1}^{K}\frac{\mu_j^2}{2\sigma_j^2}}\prod_{j=1}^{K}d\mu_j \nonumber \\
&=& \sum_i \frac{1}{(2\pi)^{K/2} n} \int \textrm{e}^{-\sum_{j=1}^{K} \frac{\mu_j^2}{2\sigma_j^2}}\prod_{j=1}^{K-1}d\mu_j \left(\sum_{j=1}^{k-1} \mu_j \sigma_j x_{ij} + \mu_k \sigma_K - y_i \right)^2 \nonumber \\
&=& \sum \frac{1}{(2\pi)^{\frac{K-1}{2}} n} \int \textrm{e}^{-\sum_{j=1}^{K-1} \frac{\mu_j^2}{2\sigma_j^2}}\prod_{j=1}^{K-1}d\mu_j \left[\sigma_K^2 + \left(\sum_{j=1}^{K-1} \mu_j \sigma_j x_{ij} - y_i\right)^2\right] \nonumber \\
&=& \sum \frac{1}{(2\pi)^{\frac{K-2}{2}} n} \int \textrm{e}^{-\sum_{j=1}^{K-2} \frac{\mu_j^2}{2\sigma_j^2}}\prod_{j=1}^{K-2}d\mu_j \left[\sigma_K^2 + \sigma_{K-1}^2 x_{i,K-1}^2 + \left(\sum_{j=1}^{K-2} \mu_j \sigma_j x_{ij} - y_i\right)^2\right] \nonumber \\
&=& ...... \nonumber \\
&=& \sum_i\frac{1}{n} \left[\sigma_K^2 + \sigma_{K-1}^2 x_{i,K-1}^2 + \sigma_{K-2}^2 x_{i, K-2}^2 + ... + \sigma_1^2 x_{i,1}^2 + y_i^2\right] \nonumber \\
&=& \sigma_K^2 + \sigma_{K-1}^2 \overline{X_{K-1}^2} + \sigma_{K-2}^2 \overline{X_{K-2}^2} + ...... + \sigma_{2}^2 \overline{X_{2}^2} + \sigma_{1}^2 \overline{X_{1}^2} +  \overline{Y^2}\label{equ_normal_energy}.
\end{eqnarray}

On the other hand, after phase transition the system reaches to its lowest energy point that
\begin{equation}
E_f= \textrm{min\;MSE} = \frac{1}{n}{\rm tr}\{C_x - C_{xy}C_y^{-1} C_{yx}\},
\end{equation}
where $C_x$ and $C_y$ are auto-covariance matrix of $x$ and $y$, respectively, and $C_{xy}$ is cross-covariance matrix between $x$ and $y$. 

Therefore, the temperature of the system is
\begin{eqnarray}
T = \frac{\sum_{j=1}^{K} \sigma_j^2 \overline{X_j^2} + \overline{Y^2} -  \overline{{\rm tr}\{C_x - C_{xy}C_y^{-1} C_{yx}\}}}{\sum_{j=1}^K \ln \sigma_j + \frac{K}{2}[1+ \ln(2 \pi)]}\label{equ_gauss_mse_highD},
\end{eqnarray}
where we denote $x_{iK} = 1$ and $\overline{X_K^2} =1$. 

For the asymptotic case  $\{\sigma_j\}\rightarrow \sigma \rightarrow \infty$, we obtain the temperature
\begin{eqnarray}
T &\sim& \frac{\sigma^2 \sum_{j=1}^{K}\overline{X_j^2}}{K \ln \sigma} \sim \left(\frac{\sum_{j=1}^{K} \overline{X_j^2}}{K}\right)\left(\frac{\sigma^2}{\ln \sigma}\right) \nonumber \\
&\sim & \overline{X^2} \left(\frac{\sigma^2}{\ln \sigma}\right)\label{equ_MSE_temp2}.
\end{eqnarray}
This result combined with  Section \ref{section_MSE_normal_2D} shows that the temperature of a ML system is determined by its data distribution, as well as parameter initialized structure.

\subsection{Parameter Initialization:  Uniform Distribution}\label{section_MSE_uniform}

If the parameters of the system initial distribution is a uniform distribution, where the $K$-dimensional parameters $\{\pmb \mu\}$ distribution satisfies
\begin{equation}
f({\pmb \mu}) = \prod_{k=1}^K\frac{1}{l_k}\label{equ_uniform_distr}
\end{equation}
for $\{-l_k/2 \leq \mu_k < l_k/2\}_{k=1, 2, 3, ..., K}$. Thus, the differential entropy of the system initial state is
\begin{equation}
S = \ln \left(\prod_{k=1}^{K}l_k \right).
\end{equation}

\subsubsection{2D Uniform Distribution}\label{section_MSE_uniform_2D}

In this special case, the average energy of the state can be calculated by
\begin{eqnarray}
\langle E_0 \rangle &= & \int \frac{1}{n} \sum_{i=1}^{n} (\mu_1 x_i + \mu_2 -y_i)^2 \frac{1}{l1} d\mu_1 \frac{1}{ l2} d\mu_2 \nonumber \\
&= & \int_{-l1/2}^{l1/2} \sum \frac{d \mu_1}{ n l_1 l_2} \int_{-l2/2}^{l2/2} [\mu_2 + (\mu_1 x_i - y_i)]^2 d\mu_2 \nonumber \\
&= & \sum \int_{-l_1/2}^{l_1/2} \frac{x_i^2}{n l_1}\left[\frac{l_2^2}{12 x_i^2} + \left(\mu_1 - \frac{y_i}{x_i}\right)^2\right] d\mu_1\nonumber \\
&= & \sum \frac{1}{n}\left[\frac{l_1^2 x_i^2}{12} + y_i^2 +  \frac{l_2^2}{12}\right] \nonumber \\
&= & \frac{l_1^2}{12}\overline{X^2} + \overline{Y^2} + \frac{l_2^2}{12}.
\end{eqnarray}
While the final state $E_f$  follows Equation (\ref{equ_MSE_Ef}). Therefore, the temperature of the system is
\begin{eqnarray}
T =  \frac{ l_1^2 \overline{X^2} + l_2^2 + 12 \overline{Y^2} -  12(1 - \rho^2) \textrm{Var}(Y)}{12 \ln (l_1 l_2)}.
\end{eqnarray}
We can show that as $l_1 = l_2 = l \rightarrow \infty$, the asymptotic temperature of the system is
\begin{equation}
T \sim \frac{l^2 (1+ \overline{X^2}) +12 \overline{Y^2}}{12 \ln l^2} \sim \left(\frac{l^2}{24\ln  l}\right) (1+ \overline{X^2})\label{equ_T_MSE_uniform}.
\end{equation}

We can compare the temperature in Equation (\ref{equ_T_MSE_uniform}) for a system with parameters initialized by uniform distributions (hereafter ``Uniform System'' with Type I State temperature $T_{\rm uniform}$), to the temperature in Equation (\ref{equ_T_MSE_normal}) for a system with parameters initialized by normal distributions (hereafter ``Normal System'' with Type I State temperature $T_{\rm normal}$). We cut out a $l^K$ sized cube in the parameter space of the Normal System, where $l$ corresponds to the distribution region in the parameter space of System Uniform. If $l=4 \sigma$, the cube encompasses approximately $0.9545^K$ probability of all particles in the Normal System. Therefore, we require $l \geq 4\sigma$ to ensure that the Uniform System has at least a $\sim 0.9545$ similarity to the Normal System. Consequently, the two temperatures under this constraint have
\begin{eqnarray}
T_{\rm uniform} &\sim& \left(\frac{l^2}{24\ln  l}\right) (1+ \overline{X^2}) \gtrsim \left(\frac{16\sigma^2}{24\ln  \sigma}\right) (1+ \overline{X^2}) \nonumber \\
&>& \left(\frac{\sigma^2}{2\ln  \sigma}\right) (1+ \overline{X^2}) \sim T_{\rm normal},
\end{eqnarray}
which means for the Normal System and the Uniform System which are similar,  the Uniform System has a higher temperature $T_{\rm uniform} > T_{\rm normal}$.

\subsubsection{High-dimensional Distribution}\label{section_MSE_uniform_highD}

Generally, assuming the initial parameters of the ML system follow a high-dimensional uniform distribution as shown in Equation (\ref{equ_uniform_distr}), we can calculate the initial energy of this system as

\begin{eqnarray}
\langle E_0 \rangle &= &\int \frac{1}{n} \sum_i \left(\sum_{j=1}^{K-1} \mu_j x_{ij} + \mu_K - y_i \right)^2 \prod_{j=1}^{K}\left(\frac{d\mu_j}{l_j}\right) \nonumber \\
&= & \sum_i \frac{1}{n} \int \prod_{j=1}^{K-1} \left(\frac{d\mu_j}{l_j}\right) \left[\frac{l_K^2}{12} +  \left(\sum_{j=1}^{K-1} \mu_j x_{ij} - y_i\right)^2\right]  \nonumber \\
&=& \frac{l_K^2}{12} + \frac{l_{K-1}^2}{12}\overline{X_{K-1}^2} +  \sum_i \frac{1}{n}  \int\prod_{j=1}^{K-3} \left(\frac{d\mu_j}{l_j}\right) \left[ x_{i,K-2}^2 \frac{l_{K-2}^2}{12} + \left(\sum_{j=1}^{K-3} \mu_j x_{ij} - y_i\right)^2\right] \nonumber \\
&=&....... =  \frac{l_K^2}{12} + \frac{l_{K-1}^2}{12}\overline{X_{K-1}^2} + ... +  \frac{l_{2}^2}{12}\overline{X_{2}^2} + \sum_i \frac{1}{n} \int_{-l_1/2}^{l_1/2} (\mu_1 x_{i1} - y_i)^2 d\mu_1 \nonumber \\
&=& \frac{l_K^2}{12} + \frac{l_{K-1}^2}{12}\overline{X_{K-1}^2} + ... +  \frac{l_{2}^2}{12}\overline{X_{2}^2}  +  \frac{l_{1}^2}{12}\overline{X_1^2} + \overline{Y^2} \nonumber \\
&=& \frac{1}{12} \sum_{j=1}^K l_j^2 \overline{X_j^2} + \overline{Y^2}
\end{eqnarray}
where we assume the $k$-th component of $\pmb x$ that $x_{iK} = 1$ always holds, so $\overline{X_K^2} =1$.  Thus, the temperature of the system is
\begin{eqnarray}
T &=& \frac{\sum_{j=1}^{K} l_j^2 \overline{X_j^2}/12 + \overline{Y}^2 -  {\rm tr}\{C_x - C_{xy}C_y^{-1} C_{yx}\}}{\sum_{j=1}^K \ln l_j} .
\end{eqnarray}
For the asymptotic case $\{l_j\} \rightarrow l \rightarrow \infty$, we have
\begin{eqnarray}
T \sim \frac{l^2 \sum_{1}^{K}\overline{X_j^2}}{12 K \ln l} \sim \left(\frac{l^2}{12 \ln l}\right)\overline{X^2}.
\end{eqnarray}

For two similar Normal and Uniform Systems with finite region $l \geq 4\sigma$, we obtain $T_{\rm uniform} > T_{\rm normal}$.

\subsection{Mixed Distribution}\label{section_MSE_mixed}

In general, the initial parameters of a system typically follow either a uniform distribution or a normal distribution. If we consider the most common system, where the distribution of the initial parameters is a mixture of uniform and normal distribution, suppose the parameter space is of dimension $K+Q$ with the first $K$ dimensions being normal distribution, denoted as $\{\mu_j^{(n)}\}_{j = 1, 2, ...,K}$ with $f_n (\mu_j)$ being normal distribution with standard deviation $\sigma_j$, and dimensions from $K+1$st to $K+Q-$th being uniform distributions, denoted as $\{\mu_j^{(u)}\}_{j = K+1, ...,K+Q}$ with $f_u(\mu_j) = 1/l_{j}$ for $-l_{j}/2 \leq \mu_j \leq l_{j}/2$  , then the energy of the system can be written as
\begin{eqnarray}
\langle E_0 \rangle &=& \frac{1}{n}\sum_i \int\left(\sum_{j=1}^{K}\mu_j^{(n)}x_{ij} + \sum_{k=K+1}^{K+Q-1}\mu_{j}^{(u)}x_{ij} + \mu_{K+Q} - y_i\right)^2 \prod_{j=1}^{K}[ f_n (\mu_j^{(n)})d\mu_j]\prod_{j=K+1}^{K+Q}[ f_u (\mu_j^{(u)})d\mu_j] \nonumber \\
&=& \frac{l_{K+Q}^2}{12} +  \frac{l_{K+Q-1}^2}{12}\overline{X_{K+Q-1}^2} + ... +  \frac{l_{K+1}^2}{12}\overline{X_{K+1}^2}  +   \frac{1}{n}\sum_i \int\left(\sum_{j=1}^{K}\mu_j^{(n)}x_{ij} - y_i\right)^2 \prod_{j=1}^{K}[ f_n (\mu_j^{(n)})d\mu_j] \nonumber \\
& = & \frac{l_{K+Q}^2}{12} +  \frac{l_{K+Q-1}^2}{12}\overline{X_{K+Q-1}^2} + ... +  \frac{l_{K+1}^2}{12}\overline{X_{K+1}^2} + \sigma_{K}^2 \overline{X_{K}^2} + \sigma_{K-1}^2 \overline{X_{K-1}^2} + ...  + \sigma_{1}^2 \overline{X_{1}^2} +  \overline{Y^2} \nonumber \\
& = & \sum_{j=1}^{K} \sigma_j^2 \overline{X_j^2} + \sum_{j=K+1}^{K+Q} \frac{l_j^2}{12} \overline{X_j^2} +  \overline{Y^2}.
\end{eqnarray}
Let we denote $\sigma_j = l_j/\sqrt{12}$, we have
\begin{eqnarray}
\langle E_0 \rangle &=& \sum_{j=1}^{K} \sigma_j^2 \overline{X_j^2} + \sum_{j=K+1}^{K+Q} \sigma_j^2 \overline{X_j^2} +  \overline{Y^2} = \sum_{j=1}^{K+Q}\sigma_j^2 \overline{X_j^2} +  \overline{Y^2},
\end{eqnarray}
which shows in the same the unified form as Equation (\ref{equ_normal_energy}). 

On the other hand, the initial entropy of the system is
\begin{eqnarray}
S & = & \ln (\sigma_1 \sigma_2 .. \sigma_K) + \frac{K}{2} \ln[1+ \ln(2\pi)] + \ln (l_{K+1} l_{K+2} ... l_{K+Q}) \nonumber \\ 
& = & \ln (\sigma_1 \sigma_2 .. \sigma_K)  + \frac{K}{2} \ln[1+ \ln(2\pi)] +  \ln (\sigma_{K+1} \sigma_{K+2} ... \sigma_{K+Q}) + \frac{Q}{2}\ln 12\nonumber \\ 
& = & \ln \left(\prod_j^{K+Q} \sigma_j \right) + \frac{K}{2} \ln[1+ \ln(2\pi)] +  \frac{Q}{2}\ln 12.
\end{eqnarray}
Thus, the temperature of the system is
\begin{equation}
T = \frac{\sum_{j=1}^{K+Q}\sigma_j^2 \overline{X_j^2} +  \overline{Y^2} + {\rm const1.}}{\ln \left(\prod_j^{K+Q} \sigma_j \right) + {\rm const2}.}
\end{equation}

(1) For all $\sigma_j \rightarrow \sigma \rightarrow \infty$, 
\begin{eqnarray}
T &\approx& \frac{\sigma^2 \sum_{1}^{K+Q}\overline{X_j^2}}{(K+Q) \ln \sigma} \approx \overline{X^2} \left(\frac{\sigma^2}{\ln \sigma}\right)\label{equ_mixed_temp1},
\end{eqnarray}
which is the same as normal distribution. 

(2) As discussed in Section \ref{section_MSE_normal_highD}, if we use $\sigma_j = \sigma $ only for $j=1,2,3, ..., K$ while $l_j \geq 4 \sigma$, i.e., $\sigma_j \geq 2\sigma/\sqrt{3}$ for $j=K+1, K+2, ..., K+Q$ , the temperature
\begin{eqnarray}
T &\gtrsim& \frac{\sigma^2 \sum_{1}^{K}\overline{X_j^2} + \frac{4}{3}\sigma^2 \sum_{K+1}^{K+Q}\overline{X_j^2}}{(K+Q) \ln \sigma} > \overline{X^2} \left(\frac{\sigma^2}{\ln \sigma}\right)\label{equ_mixed_temp2},
\end{eqnarray}
which gives a higher temperature.

\subsection{MSE with Regularization}\label{section_MSE_regular}

In this section, we explore the system energy with regularization, especially $L_1$ and $L_2$ regularization. For the MSE-based energy, the energy with $L_1$ regularization is $E_{L_1} = E_{\rm MSE} + \lambda \sum_j |\mu_j|$, while the energy with $L_2$ regularization is $E_{L_2} = E_{\rm MSE} + \lambda \sum_j |\mu_j^2|$. 

\subsubsection{Normal Distribution}

If the initial distribution of the ML system parameters follow normal distributions, then the initial energy of the system with $L_1$ regularization is calculated as
\begin{eqnarray}
\langle E_{L_1}\rangle & = & \langle E_{0}\rangle + \lambda \sum_{j=1}^K \left(\int \frac{|\mu_j|}{\sqrt{2\pi} \sigma_j} \textrm{e}^{-\frac{\mu_j^2}{2\sigma_j^2}}d \mu_j\right) \nonumber \\ 
&= & \langle E_{0}\rangle + \lambda\sqrt{\frac{2}{\pi}} \sum_{j=1}^{K}\sigma_j = \sum_{j=1}^K \sigma_j (\sigma_j \overline{X_j^2} + \lambda') + \overline{Y^2},
\end{eqnarray}
where we take $\lambda' = \lambda\sqrt{2/\pi}$. 

The system with $L_2$ regularization is 
\begin{eqnarray}
\langle E_{L_1}\rangle & = & \langle E_{0}\rangle + \lambda \sum_{j=1}^K \left(\int \frac{|\mu_j^2|}{\sqrt{2\pi} \sigma_j} \textrm{e}^{-\frac{\mu_j^2}{2\sigma_j^2}}d \mu_j\right) \nonumber \\ 
&= & \langle E_{0}\rangle + \lambda \sum_{j=1}^{K} \sigma_j^2 = \sum_{j=1}^K \sigma^2_j \left(\overline{X_j^2} + \lambda \right) + \overline{Y^2}.
\end{eqnarray}

For the asymptotic case all $\sigma_j \rightarrow \sigma \rightarrow \infty$, we obtain two temperature forms
\begin{equation}
T_{L_1} \sim \overline{X^2}\left(\frac{\sigma^2}{\ln \sigma}\right) \sim T_{\rm normal},
\end{equation}
\begin{equation}
T_{L_2} \sim (\overline{X^2} + \lambda) \left(\frac{\sigma^2}{\ln \sigma}\right) \sim \left(1 + \frac{\lambda}{\overline{X^2}}\right) T_{\rm normal}.
\end{equation}
This implies that $L_1$ regularization, in the asymptotic sense, does not change the system temperature, while $L_2$ regularization maintains the temperature form ($\propto \frac{\sigma^2}{\ln \sigma}$),  but adds $(\lambda/\overline{X^2}) T_{\rm normal} $ to the existing system temperature.


\subsubsection{Uniform Distribution}

Similar result can be obtained for initially uniform distributed parameters. The initial energy with  $L_1$ and $L_2$ regularization are
\begin{equation}
\langle E_{L_1}\rangle = \sum_{j=1}^{K} \frac{l_j}{12}(l_j \overline{X_j^2} + 3\lambda) + \overline{Y^2},
\end{equation}
\begin{equation}
\langle E_{L_2}\rangle = \sum_{j=1}^{K} \frac{l_j^2}{12}(\overline{X_j^2} + \lambda ) + \overline{Y^2},
\end{equation}
respectively. So for the asymptotic case all $l_j = l \rightarrow \infty$, the two temperatures are
\begin{equation}
T_{L_1} \sim \overline{X^2} \left(\frac{l^2}{12\ln l}\right) \sim T_{\rm uniform},
\end{equation}
\begin{equation}
T_{L_2} \sim (\overline{X^2} + \lambda) \left(\frac{l^2}{12\ln l}\right) \sim \left(1 + \frac{\lambda}{\overline{X^2}}\right) T_{\rm uniform}.
\end{equation}
Thus we have the similar result that $L_1$ regularization does not change the temperature, while $L_2$ regularization also does not change the energy form $\propto \frac{l^2}{12 \ln l}$.

\subsection{Linear Regression with MAE}\label{section_MAE_linear}

Also, we consider systems with energy follow the MAE form that $E_p = \frac{1}{n}|\sum_j^{K-1} \mu_j x_{ij} + \mu_K - y_i|$. For initialized normal distributed parameters, the initial energy of the state is
\begin{eqnarray}
\langle E_0 \rangle &= & \int \frac{1}{n} \sum_{i=1}^n \left|\sum_j^{K-1} \mu_j x_{ij} + \mu_K - y_i \right| \prod_{j=1}^K \left(\frac{1}{\sqrt{2 \pi} \sigma_j} \textrm{e}^{-\frac{\mu_j^2}{2 \sigma_j^2}} d\mu_j\right)\label{equ_MAE_energy1}
\end{eqnarray}
Different from MSE, the energy in the form of MAE does not have an analytical solution. We look into the asymptotic properties of the energy. Assuming $\sigma_1 \gg \{\sigma_2, \sigma_3, ..., \sigma_K\}$, Equation (\ref{equ_MAE_energy1}) becomes
\begin{eqnarray}
\langle E_0 \rangle \sim \int \frac{1}{\sqrt{2 \pi} n} \sum_{i=1}^n |\mu_1 \sigma_1 x_{i1}| \textrm{e}^{-\frac{\mu_1^2}{2}}d\mu_1 \sim \sqrt{\frac{2}{\pi}} \sigma_1 \overline{|x_1|},
\end{eqnarray}
where $\overline{|x_1|}$ is the average of absolute values of the first component of $\pmb x$.  Therefore, for $\{\sigma_j\}_{j=1, 2, ..., K}\rightarrow\infty$, the asymptotic energy is approximated as 
\begin{equation}
\langle E_0 \rangle \sim \sqrt{\frac{2}{\pi}} \sum_{j=1} ^K \sigma_j \overline{|x_j|}.
\end{equation}
The temperature of the system is
\begin{equation}
T \sim \frac{\sqrt{\frac{2}{\pi}} \sum_{j=1} ^K \sigma_j \overline{|x_j|}}{\sum_{j=1}^K \ln \sigma_j + \frac{K}{2}[1+ \ln (2\pi)]}.
\end{equation}
For $\{\sigma_j\}_{j=1, 2, ..., K} \rightarrow \sigma \rightarrow \infty$, we have
\begin{equation}
T \sim \sqrt{\frac{2}{\pi}} \left(\frac{\sigma}{\ln \sigma}\right)\frac{\sum_{j=1} ^K \overline{|x_j|}}{K} \sim \sqrt{\frac{2}{\pi}}  \left(\frac{\sigma}{\ln \sigma}\right) \overline{|X|}.
\end{equation}

According to Section \ref{section_MSE_normal_highD}, the temperature of a system under the same conditions but with MSE energy is $T_{\rm MSE} \sim \overline{X^2} \left(\frac{\sigma^2}{\ln \sigma}\right) \gg T_{\rm MAE} \sim  \sqrt{\frac{2}{\pi}}  \left(\frac{\sigma}{\ln \sigma}\right) \overline{|X|}$.  Since $T_{\rm MSE} \propto \sigma^2$ while $T_{\rm MAE} \propto \sigma$, system with MSE energy has much higher temperature than that with MAE energy. 

For the case where the initial distribution of the parameters is uniform, we can calculate the analytical solution for $\langle E_0\rangle$.  For example, for the two-dimensional case:

\begin{eqnarray}
\langle E_0 \rangle = \int \frac{1}{n} \sum_i |\mu_1 x_i + \mu_2 - y_i| \frac{d\mu_1}{l_1} \frac{d\mu_2}{l_2},
\end{eqnarray}
we derive 
\begin{eqnarray}
6n l_1 l_2 x_i \langle E_0 \rangle &= & \sum \left(\frac{l_1}{2}x_i + \frac{l_2}{2} - y_i\right)^2 \left|\frac{l_1}{2} x_i + \frac{l_2}{2} - y_i\right|  \nonumber \\
&-& \sum \left(\frac{l_1}{2}x_i - \frac{l_2}{2} + y_i\right)^2 \left|\frac{l_1}{2} x_i - \frac{l_2}{2} + y_i\right|   \nonumber \\
&- & \sum \left(\frac{l_1}{2}x_i - \frac{l_2}{2} - y_i\right)^2 \left|\frac{l_1}{2} x_i - \frac{l_2}{2} - y_i\right| \nonumber \\
&+& \sum \left(\frac{l_1}{2}x_i + \frac{l_2}{2} + y_i\right)^2 \left|\frac{l_1}{2} x_i + \frac{l_2}{2} + y_i\right|\label{equ_MAE_energy_2}.
\end{eqnarray}
For $l_1 \gg l_2$, from Equation \ref{equ_MAE_energy_2} we can obtain $\langle E_0 \rangle \sim \frac{1}{4}l_1\overline{|x_i|}$. Similarly, for $l_2 \gg l_1$, we obtain $\langle E_0 \rangle \sim \frac{1}{4}l_2$.

For high dimensions, 
\begin{eqnarray}
 \langle E_0 \rangle &= & \int\frac{1}{n}\sum_i \left|\sum_{j=2}^{K-1} \mu_j x_{ij} + \mu_K - y_i\right| \prod_{j=1}^{K} \left(\frac{d \mu_j}{l_j}\right).
\end{eqnarray}
We can get the asymptotic solution of the energy
\begin{equation}
 \langle E_0 \rangle  \sim \frac{1}{4}[l_1 |x_{i1}| + l_2 |x_{i2}| + l_3 |x_{i3}| + ... + l_{k_1}|x_{i, K-1}| + l_K],
\end{equation}
and the temperature is
\begin{equation}
 T  \sim \frac{\sum_{j=1}^{K} l_j |x_{ij}|}{4 \sum_{j=1}^{K} \ln l_j}.
 \end{equation}
 For $\{l_i\} \rightarrow l \rightarrow \infty$, we have 
 \begin{equation}
T \sim \frac{1}{4} \left(\frac{l}{\ln l }\right) |\overline{X}|.
 \end{equation}
 Compare systems under same conditions but with different energy forms, 
 \begin{equation}
 T_{\rm MAE} \propto \frac{l}{\ln l} \ll T_{\rm MSE} \propto \frac{l^2}{\ln l}. 
 \end{equation}
Again, we find that MSE gives much higher temperature than MAE.

\subsection{Physical Explanation of Temperature}\label{section_physical_explan}

Let us consider two ML systems $A$ and $B$. For simplification, we assume the systems have a linear regression model structure and MSE energy form, and the initial normal distribution of the parameters is determined by a sufficiently large $\sigma$\footnote{For the system with initial uniform distribution of parameters, we can use a sufficiently large $l$  to derive very similar results in this section. }. Thus, the temperature of the system can be calculated by Equation (\ref{equ_MSE_temp2}), which means the system temperature is determined by data distribution $X$ and the initial $\sigma$ from the parameter space. 

\begin{figure}[h]
    \centering
    \begin{subfigure}[b]{0.55\textwidth}
        \includegraphics[width=\textwidth]{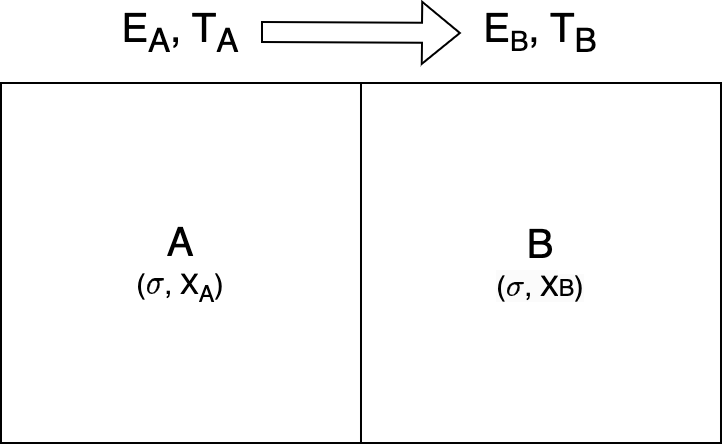}
    \end{subfigure}
    \hfill
    \begin{subfigure}[b]{0.35\textwidth}
        \includegraphics[width=\textwidth]{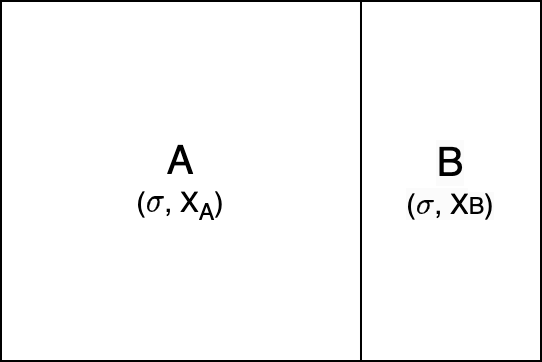}
    \end{subfigure}
\caption{\small ML systems: mixing (left panel) and training (right panel).  For two systems $A$ and $B$ that have not yet reached equilibrium, suppose they have the same parameter distribution ($\sigma$), but different temperatures and energies $(T_A,E_A)$ and $(T_B, E_B)$ respectively. If $T_A > T_B$, then after mixing, the system's temperature $T_{A \cup B}$ is between $T_A$ and $T_B$, with the direction of energy flow from $A$ to $B$, meaning the energy of system $A$ decreases while the energy of system $B$ increases. This process by which systems $A$ and $B$ reach equilibrium can be analogized to a physical process in thermodynamic systems, with very strong similarities. On the other hand, if the data of a system changes from $A$ to $C$ and needs to be retrained, we can view the training process as the mixing process of systems $A$ and $B$, where $B = C-A$.}\label{fig_temp_systems}
\end{figure}

\paragraph{Equilibrium Systems}

From the above assumption we can obtain the temperatures of the two systems: 
\begin{equation}
T_{A} \sim \frac{\sigma^2}{\ln \sigma}\overline{X_A^2},\quad T_{B} \sim \frac{\sigma^2}{\ln \sigma}\overline{X_B^2},
\end{equation}
If $X_A$ and $X_B$ have the same data distribution, i.e., $\overline{X_A^2} = \overline{X_B^2}$, we can derive $T_A = T_B$. 

Next we mix systems $A$ and $B$ to form a new system $A \cup B$,  i.e., the data of the two systems are combined. Since $\overline{X_A^2} = \overline{X_B^2} = \overline {X_{A \cup B}^2}$, we get the temperature of the mixed system that
\begin{equation}
T_{A \cup B} \sim \frac{\sigma^2}{\ln \sigma} X_{A \cup B}^2 = T_A = T_B. 
\end{equation}
From a physical perspective, systems $A$ and $B$ are in equilibrium with the same temperature.  Therefore, systems with the same temperature still have the same temperature after mixture. 

\paragraph{Non-Equilibrium Systems} 

Now let us image that the data distributions of systems $A$ and $B$ are significantly different, such as 
\begin{eqnarray}
&& x > 1 \quad \textrm{for}  \quad \forall x \in X_A \\
&& x < 1 \quad \textrm{for}  \quad \forall x \in X_B.
\end{eqnarray}
Thus, the temperatures of the two systems are
\begin{equation}
T_A \sim \frac{\sigma^2}{\ln \sigma}\overline{X_A^2} > T_B \sim \frac{\sigma^2}{\ln \sigma}\overline{X_B^2}.
\end{equation}
For the mixed system $A \cup B$, since $\overline{X_B^2} < \overline{X_{A \cup B}^2} < \overline{X_A^2}$, we can get
\begin{equation}
T_B < T_{A \cup B} \sim \frac{\sigma^2}{\ln \sigma} \overline{X_{A \cup B}^2} < T_A.
\end{equation}
The temperature of the mixed  system $A \cup B$ is between the two pre-mixed systems. 

Note that the energy of systems $A$ and $B$ are
\begin{eqnarray}
E_A \sim K_A \overline{X_A^2}, \quad E_B \sim K_B \overline{X_B^2},
\end{eqnarray}
where $K_A$ and $K_B$ are the dimensions of parameter space in $A$ and $B$, respectively. The mixed system redistributed energy as
\begin{eqnarray}
E_A' \sim K_A  \overline{X_{A \cup B}^2}, \quad E_B' \sim K_B  \overline{X_{A \cup B}^2}.
\end{eqnarray}
For $T_A \geq T_B$, the energy changes in $A$ and $B$ are 
\begin{eqnarray}
\Delta E_A &=& K_A \sigma^2 (\overline{X_{A \cup B}^2} - \overline{X_{A}^2}) < 0, \\
\Delta E_B &=& K_B \sigma^2 (\overline{X_{A \cup B}^2} - \overline{X_{B}^2}) > 0.
\end{eqnarray}
The energy is transferred from system $A$ to system $B$. In other words, the energy flows from a high-temperature, high-energy system to a low-temperature, low-energy system once they reach equilibrium. This phenomenon aligns with the behavior of physical systems in a thermodynamic context.

\paragraph{Physical Scenario of Model Retraining}

Now, let us consider a ML system with data $X_A$. The system obtains new data $X_B$ with a different distribution that  $\overline{X_A^2} \neq \overline{X_B^2}$. We mix $A$ and $B$ together to form a new system $A \cup B$ and retrain the model. Since two non-equilibrium systems are mixed, the temperature $T_{A \cup B}$ of the new system has the following characteristics: If $T_A < T_B$, we have $T_A < T_{A \cup B} < T_B$, else if $T_A > T_B$, we have $T_A > T_{A \cup B} > T_B$.

That is to say, if the new data added to the ML system has a higher temperature, then the temperature of the system after retraining will also increase, and vice versa. The scenarios discussed above temperature changes in equilibrium, non-equilibrium, and model retraining systems are very similar to traditional thermodynamic systems. This is the thermodynamic landscape of the temperature in ML systems.

\section{Logistic Regression with Cross Entropy}\label{section_LR_CE}

\subsection{Parameter Initialization: Normal Distribution }\label{section_CE_normal}

We move forward to discuss ML systems with cross entropy energy, which is written as
\begin{eqnarray}
E_p &=& - \sum_i \frac{1}{n} [y_i \ln \hat{y_i} + (1-y_i) \ln (1- \hat{y}_i)] \nonumber \\
&=& \sum_i \frac{1}{n}[\ln (1+ \textrm{e}^{-z_i})  + (1-y_i)z_i],
\end{eqnarray}
given $z_i = \sum_{j=1}^{K-1}\mu_j x_{ij} + \mu_K$. 

Note that 
\begin{equation}
\int z_i f({\pmb \mu})d{\pmb \mu} =0
\end{equation}
if the distribution function $f({\pmb \mu})$ is an even function. Thus, the average energy of the initial state with parameter normal distribution is
\begin{eqnarray}
\langle E_0 \rangle &=& \int E_p f({\pmb \mu}) d{\pmb \mu} = \int \sum_i \frac{1}{n} \ln (1+ \textrm{e}^{-z_i}) \prod_j \left(\frac{1}{\sqrt{2 \pi} \sigma_j} \textrm{e}^{-\frac{\mu_j^2}{2\sigma_j^2}}d\mu_j\right) \nonumber \\
&= & \int \sum_i \frac{1}{n} \ln \left[1 + \exp\left(-\sum_{j=1}^{K-1} \mu_j \sigma_j x_{ij} - \mu_K \sigma_K\right)\right] \prod_j \left(\frac{1}{\sqrt{2 \pi}} \textrm{e}^{-\frac{\mu_j^2}{2}}d\mu_j\right).\label{equ_ce_E0_1}
\end{eqnarray}
Since $\langle E_0\rangle$ does not have an analytic solution, we examine the asymptotic solution of $\langle E_0\rangle$ for any   $\sigma_j$ in  $\{\sigma_1, \sigma_2, ... \sigma_K\} \rightarrow \infty$. We discuss this by considering two cases: Case I and Case II as follows.

\paragraph{Case I}:  $z_i' = \sum_{j=1}^{K-1} \mu_j \sigma_j x_{ij} + \mu_K \sigma_K \rightarrow \infty$. Thus, we have $\ln (1 + \textrm{e}^{-z_i'}) \rightarrow \textrm{e}^{-z_i'}$. From Equation (\ref{equ_ce_E0_1}) we can derive
\begin{eqnarray}
\langle E_0 \rangle_i &\sim& \int \frac{1}{n} \textrm{e}^{-z_i'} \prod_j \left(\frac{1}{\sqrt{2 \pi}} \textrm{e}^{-\frac{\mu_j^2}{2}}d\mu_j\right) \nonumber \\
& \sim & \int \frac{1}{\sqrt{2 \pi}} \prod_{j=1}^{K-1} \exp\left[-\mu_j \sigma_j x_{ij} - \frac{\mu_j^2}{2}\right]d\mu_j \exp\left[-\mu_K \sigma_K - \frac{\mu_K^2}{2}\right]d\mu_K \nonumber \\
& \sim & \frac{1}{n} \exp\left[\frac{1}{2} \sum_{j=1}^{K-1}\sigma_j^2 x_{ij}^2 + \frac{1}{2} \sigma_K^2\right]\label{equ_ce_E0_2}.
\end{eqnarray}

\paragraph{Case II}:  $z_i' = \sum_{j=1}^{K-1} \mu_j \sigma_j x_{ij} + \mu_K \sigma_K \rightarrow -\infty$. In this case, we have $\ln (1 + \textrm{e}^{-z_i'}) = \ln [\textrm{e}^{-z_i'}(1 + \textrm{e}^{z_i'})] = -z_i' + \ln(1 + \textrm {e}^{z_i'}) \rightarrow -z_i' + \textrm{e}^{z_i'}$. Note that $\int z_i' f({\pmb \mu})d{\pmb \mu} =0$ also  holds, so Equation (\ref{equ_ce_E0_1}) becomes
\begin{eqnarray}
\langle E_0 \rangle &\sim& \int  \frac{1}{n} \exp\left[\sum_{j=1}^{K-1} \mu_j \sigma_j x_{ij} + \mu_K \sigma_K \right]  \prod_j \left(\frac{1}{\sqrt{2 \pi}} \textrm{e}^{-\frac{\mu_j^2}{2}}d\mu_j\right) \nonumber \\
 & \sim & \frac{1}{n} \exp\left[\frac{1}{2} \sum_{j=1}^{K-1}\sigma_j^2 x_{ij}^2 + \frac{1}{2} \sigma_K^2\right],
\end{eqnarray}
which is exactly the same as Equation (\ref{equ_ce_E0_2}). So we can write 
\begin{eqnarray}
\langle E_0 \rangle &\sim& \exp\left[\frac{\sigma_K^2}{2}\right]\overline{\exp\left[\frac{1}{2} \sum_{j=1}^{K-1}\sigma_j^2 x_{ij}^2\right]} \nonumber \\
&> & \textrm{e}^{\sigma_K^2/2} \exp\left[\frac{1}{2} \sigma_j^2 (x_{j}^{\rm min})^2\right] \\
&> & \exp\left[\frac{1}{2} \sigma_j^2 (x_{j}^{\rm min})^2\right],
\end{eqnarray}
where $x_{j}^{\rm min} = {\rm min}\{x_{1j}, x_{2j}, ... x_{nj}\}$. Therefore, the temperature of the system is 
\begin{eqnarray}
T \sim \frac{\langle E_0\rangle - E_{f}}{\sum_j \ln \sigma_j  + \frac{K}{2}[1 + \ln (2\pi)]} > \frac{\textrm {e}^{\sigma_j^2 (x_j^{\rm min})^2/2} - E_f}{\sum_j \ln \sigma_j  + \frac{K}{2}[1 + \ln (2\pi)]}.
\end{eqnarray}
If we have only one parameter $\sigma_j\in \{\sigma_1, \sigma_2, ... \sigma_K\} \rightarrow \infty$, the system temperature is
\begin{equation}
T \gtrsim \frac{\textrm {e}^{\sigma_j^2 (x_j^{\rm min})^2/2} - E_f}{\ln \sigma_j} \sim \frac{\textrm{e}^{\sigma_j^2 (x_j^{\rm min})^2/2}}{\ln \sigma_j}.
\end{equation}
For all parameters $\{\sigma_1, \sigma_2, ..., \sigma_K\} \rightarrow \sigma \rightarrow \infty$, we obtain the temperature
\begin{eqnarray}
   T &\sim& \frac{\exp\left[\frac{\sigma^2}{2}\right]\overline{\exp\left[\frac{1}{2} \sum_{j=1}^{K-1}\sigma^2 x_{ij}^2\right]} - E_f}{K \ln \sigma} \nonumber \\
   &>& \frac{\textrm{e}^{\sigma^2/2}}{K \ln \sigma} \propto \frac{\textrm{e}^{\sigma^2/2}}{\ln \sigma}. 
\end{eqnarray}
Note that for MSE and MAE energy the asymptotic temperature is $\propto \sigma^2/(\ln \sigma)$ and $\sigma/(\ln \sigma)$ respectively, so the temperature from system with cross entropy energy $T_{\rm C.E.} \gg T_{\rm MSE}, T_{\rm MAE}$.

\subsection{Parameter Initialization:  Uniform Distribution}\label{section_CE_uniform}

For parameters initialized by uniform distribution, the initial average energy is written as 
\begin{eqnarray}
\langle E_0 \rangle = \int \sum_i \frac{1}{n} \left[1 + \exp\left(-\sum_{j=1}^{K-1} \mu_j l_j x_{ij} - \mu_K l_K \right)\right] \prod_{j=1}^{K} d \mu_j\label{equ_ce_E0_3}.
\end{eqnarray}
Given the asymptotic case $\sum_{j=1}^{K-1} \mu_j l_j x_{ij} - \mu_K l_K \rightarrow \infty$ (Case I), Equation (\ref{equ_ce_E0_3})  becomes
\begin{eqnarray}
\langle E_0 \rangle &\sim& \sum_i \frac{1}{n} \int_{-1/2}^{1/2}\exp\left(-\sum_{j=1}^{K-1} \mu_j l_j x_{ij}\right) d\mu_j \int_{-1/2}^{1/2}\exp\left(-\mu_K l_K \right)d\mu_K \nonumber \\
&\sim& \sum_i \frac{1}{n} \prod_{j=1}^{K-1}\left( \frac{\textrm{e}^{l_j |x_{ij}|/2} - \textrm{e}^{-l_j |x_{ij}|/2}}{l_j |x_{ij}|}\right)\left( \frac{\textrm{e}^{l_K/2} - \textrm{e}^{-l_K/2}}{l_K}\right) \nonumber \\
&\sim & \frac{1}{n} \sum_i \prod_{j=1}^{K-1} \left(\frac{\textrm{e}^{l_j |x_{ij}|/2}}{l_j |x_{ij}|}\right) \left(\frac{\textrm{e}^{l_K/2}}{l_K}\right)\label{equ_ce_E0_4}.
\end{eqnarray}
Note that we assume $\{l_1, l_2, ... l_K\} \rightarrow \infty$ for the last step derivation in Equation \ref{equ_ce_E0_4}. On the other hand, for another asymptotic case $\sum_{j=1}^{K-1} \mu_j l_j x_{ij} - \mu_K l_K \rightarrow -\infty$ (Case II), we can check that Equation (\ref{equ_ce_E0_4}) still holds. Also note that the function $e^{x/2}/x$ has a minimum value $e/2$. For $\{l_1, l_2, ... l_K\} \rightarrow l  \rightarrow \infty$, the energy shows
\begin{eqnarray}
\langle E_0 \rangle > \frac{1}{n} \sum_i \left(\frac{e}{2}\right)^{K-1} \left(\frac{e^{l/2}}{l}\right) = \left(\frac{e}{2}\right)^{K-1} \left(\frac{e^{l/2}}{l}\right).
\end{eqnarray}
Thus, the temperature of the system is
\begin{eqnarray}
T = \frac{\langle E_0\rangle - E_f}{K \ln l} \gtrsim  \left(\frac{e}{2}\right)^{K-1} \frac{e^{l/2}}{K l \ln l} > 0.614 \frac{e^{l/2}}{l \ln l} \propto \frac{e^{l/2}}{l \ln l}
\end{eqnarray}
Compared with temperature in system with MSE or MAE energy that $T_{\rm MSE} \propto l^2/\ln l$ and $T_{\rm MAE} \propto l/\ln l$, we get the conclusion that $T_{\rm C.E.} \gg T_{\rm MSE}, T_{\rm MAE}$.

\section{Temperature in Neural Network}\label{section_nn}

In this section, we look into a standard artificial neural network system. We assume the neural network has $L$ layers. For the $p$-th layer, the input data is  ${\pmb a}^{p-1}$, the output data is ${\pmb a}^{p}$, and the size of the layer is $l_p$. We have
\begin{eqnarray}
&&{\bf z}^p = {\bf W}^p {\bf a}^{p-1} + {\bf b}^p \\
&&{\bf a}^p = \phi({\bf z}^p), 
\end{eqnarray}
where $\phi$ is the activation function. Some typical activation function including Sigmoid function $\sigma(z) = 1/(1+ \textrm{e}^{-z})$, Tanh function ${\rm tanh}(z) = (\textrm{e}^z - \textrm{e}^{-z})/(\textrm{e}^z + \textrm{e}^{-z})$ and ReLU function ReLU$(z) = {\rm max}\{0, z\}$.

\subsection{Initial Normal Distribution}\label{section_nn_normal}

\subsubsection{Default Case: Tanh Activation Function}\label{section_nn_default}

We assume the system follows MSE to set up its energy 
\begin{equation}
E =  \sum({\bf W}^L {\bf a}^{L-1} + {\bf b}^L - y)^2,
\end{equation}
with ${\bf W}^L = \{W^L_{ij}\}$ and $i \in [1, l_L]$, $j\in [1, l_{L-1}]$ being the weight matrix connecting the last two layers.

Let us consider the asymptotic solution of the energy equation above when all parameters have initially normal distributions. For any element $W_{ij}^p$ in any weight matrix in the neural network, as the asymptotic standard deviation $\sigma_{ij}^p \rightarrow \infty$, $\pmb z^p \rightarrow \pm \infty$. We first discuss the case where the activation function is ${\rm tanh}(z)$ with $|{\rm tanh}(z)| \rightarrow 1$ so that  $|{\pmb a}^p| = |{\rm tanh}(z)| \rightarrow 1$. The energy of the system is written as
\begin{eqnarray}
\langle E_0 \rangle &=& \frac{1}{n}\int \sum_{i=1}^n \sum_{k=1}^{l_L} \left(\sum_{j=1}^{l_{L-1}} W_{kj}^L a_{ij}^{L-1} + b_k^L - y_{ik}\right)^2 f({\pmb W}^L) d{\pmb W^L} db^L \nonumber \\
&=& \frac{1}{n}\sum_i \sum_k \left[\sum_j (\sigma_{kj}^L)^2 (a_{ij}^{L-1})^2 + (\sigma_k^L)^2 + (y_{ik})^2\right]\nonumber \\
&\sim& \frac{1}{n} \sum_i \sum_k \left[\sum_j (\sigma_{kj}^L)^2 + (\sigma_k^L)^2 + (y_{ik})^2\right] \label{equ_nn_E0_1}.
\end{eqnarray}
Given $\sigma_{kj}^L, \sigma_{k}^L \rightarrow \sigma \rightarrow \infty$, Equation (\ref{equ_nn_E0_1}) can be simplified to
\begin{eqnarray}
\langle E_0 \rangle &\sim& \frac{1}{n} \sum_{i,k} [\sigma^2 (l_{L-1} + 1) + y_{ik}^2] \sim \frac{1}{n} \sum_i [\sigma^2 (l_{L-1} + 1)l_L + y_i^2] \nonumber \\
&\sim & \sigma^2 (l_{L-1} + 1) l_L + \overline{Y^2} \sim \sigma^2 (l_{L-1} + 1) l_L
\end{eqnarray}
Note that the total number of parameters in the neural network is $\mathcal{D} = l_L (l_{L-1} + 1) + l_{L-1} (l_{L-2} + 1) + ... + l_1(l_0 + 1) = \sum_{p=1}^{l_L} l_p (l_{p-1} + 1)$. The temperature of the neural network system is
\begin{eqnarray}
T &=& \frac{\langle E_0 \rangle - E_f}{\sum_{1}^{\mathcal{D}}\ln \sigma + \frac{\mathcal{D}}{2}[1 + \ln (2\pi)]} \nonumber \\
&\sim& \frac{l_L (l_{L-1}+ 1)}{\sum_{p=1}^{l_L} l_p (l_{p-1} + 1)}  \left(\frac{\sigma^2}{\ln \sigma}\right)\label{equ_nn_E0_2}.
\end{eqnarray}

\subsubsection{Sigmoid Function}\label{section_nn_sigmoid}

If the activation function is the sigmoid function, i.e., $a = \sigma(z) \rightarrow \{1,0\}$ for $z \rightarrow \infty$, Equation (\ref{equ_nn_E0_1}) becomes
\begin{eqnarray}
\langle E_0 \rangle &\sim& \frac{1}{n}\sum_i \sum_j [\sigma^2 (\xi l_{L-1}+1) + y_{ik}^2] \nonumber \\
&\sim & \sigma^2 l_L (\xi l_{L-1} + 1),
\end{eqnarray}
where $\xi$ is the fraction of $a^L_{j} \rightarrow 1$. Thus, the temperature of the system is
\begin{eqnarray}
T \sim \frac{l_L (\xi l_{L-1}+ 1)}{\sum_{p=1}^{l_L} l_p (l_{p-1} + 1)}  \left(\frac{\sigma^2}{\ln \sigma}\right).  
\end{eqnarray}
Compared with Equation (\ref{equ_nn_E0_2}), the neural network with Sigmoid activation function has lower temperature than the that with tanh activation function, with an approximation $T_{\rm sigmoid} \sim \frac{\xi l_{L-1}+1}{l_{L-1} + 1} T_{\rm tanh} \sim \xi T_{\rm tanh}$.

\subsubsection{ReLU Function}\label{section_nn_relu}

If the activation function is the ReLu function, i.e., $\phi(z) = z$ for $z>0$ and $\phi(z) = 0$ for $z\leqslant 0$, for the $p$-th layer and the $j$-th component of ${\pmb z}^p$, $z_j^p = \sum_{k=1}^{l_{p-1}} W_{jk}^{p}a_k^{p-1} + b_j^p$.  For any $z_j^p >0$, we can use parameters set $ -W_{jk}^{p}$ and $-b_j^p$ to find  $\hat{z}_j^p < 0$ while $|\hat{z}_j^p| = |z_j^p|$.  From symmetry we can derive $\int_{z_j^p > 0} f({\pmb W^p}, {\pmb b^p})d{\pmb W^p} d{\pmb b^p} = \int_{z_j^p \leqslant 0} f({\pmb W^p}, {\pmb b^p})d{\pmb W^p} d{\pmb b^p}$, therefore, we have
\begin{eqnarray}
&& \int (a_j^p)^2 f({\pmb W^p}, {\pmb b^p})d{\pmb W^p} d{\pmb b^p} = \frac{1}{2}\int (z_j^p)^2 f({\pmb W^p}, {\pmb b^p})d{\pmb W^p} d{\pmb b^p} \nonumber \\
&& = \frac{1}{2} \int \left(\sum_{k=1}^{l_p} W_{jk}^{p}a_k^{p-1} + b_j^p \right)^2 f({\pmb W^p}, {\pmb b^p})d{\pmb W^p} d{\pmb b^p} \nonumber \\
&& \approx \frac{1}{2} \left[\sum_{k=1}^{l_{p-1}}(\sigma_{jk}^{p})^2 (a_{ik}^{p-1})^2 + (\sigma_j^{p})^2 \right] \nonumber \\
&& \approx \frac{1}{2} \left[\sum_{1}^{l_{p-1}} \sigma^2 (a_{ik}^{p-1})^2 + \sigma^2 \right] \label{equ_nn_E0_4}.
\end{eqnarray}

Equation (\ref{equ_nn_E0_4}) can be used to calculate the initial energy
\begin{eqnarray}
\langle E_0 \rangle &\sim& \frac{1}{n} \int \sum_i \sum_k \left[\sum_j \sigma^2 (a_{ij})^2 + \sigma^2 \right] f({\pmb W, \pmb b})d{\pmb W}d{\pmb b} \nonumber \\
&\sim& \frac{1}{2n} \sum_i \int l_L l_{L-1} (\sigma^2)^2 \left[\sum_{j= 1}^{l_{L-2}} (a_{ij}^{L-1})^2 + 1\right] f({\pmb W, \pmb b})d{\pmb W}d{\pmb b} \nonumber \\
&\sim & \frac{j = 1}{2^2 n} \sum_i \int l_L l_{L-1} l_{L-2} (\sigma^2)^3 \left[\sum_{1}^{l_{L-3}} (a_{ij}^{L-2})^2 + 1\right] f({\pmb W, \pmb b})d{\pmb W}d{\pmb b} \nonumber \\
&\sim & ... \sim \frac{1}{2^{L-1} n} \sum_i \left(\prod_{p=1}^{L} l_p\right) (\sigma^2)^{L} \left[\sum_{j=1}^{l_0} (x_{ij})^2 + 1\right] \nonumber \\
&\sim & \frac{1}{2^{L-1}} \left(\prod_{p=1}^{L} l_p\right) \sigma^{2L} \overline{X^2}\label{equ_nn_E0_5},
\end{eqnarray}
where we denote $X_i^2 = \sum_{j=1}^{l_0} x_{ij}^2 + 1$, and $\overline{X^2} = \frac{1}{n} X_i^2$. Thus, the temperature of the system is calculated as
\begin{eqnarray}
T &\sim& \frac{2^{1-L}\left(\prod_{p=1}^{L} l_p\right) \sigma^{2L} \overline{X^2}}{\sum_{p=1}^{L}l_p (l_{p-1} + 1)\ln\sigma} \nonumber \\
&\propto& \left(\frac{\sigma^{2L}}{\ln \sigma}\right) \overline{X^2},
\end{eqnarray}
which is much higher than the temperature from the system using Sigmoid or Tanh activation function that $\propto \sigma^2/\ln \sigma$. 

\subsection{Temperature in Individual Layers}\label{section_nn_layer}

From Section \ref{section_nn_default} to Section \ref{section_nn_relu}, we calculated the temperature of the entire neural network system corresponding to different activation functions. In fact, for a neural network, we can also calculate the temperature of each individual layer. To do this, we can write the MSE energy of the $p$-th layer as
\begin{equation}
({\pmb W^p} {\pmb a^{p-1}} + {\pmb b}^{p} - \hat{\pmb z^p})^2,
\end{equation}
where $\hat{\pmb z}^p$ represents the ``ideal" $\pmb z^p$ value for that layer. This value can either be back-propagated from the label $y_i$ to the $p$-th layer, or be the optimized value after the neural network training 
converged. In any case, under asymptotic condition, we have $|{\pmb W^p} {\pmb a^{p-1}}|$, $|{\pmb b}^{p}| \gg \hat{\pmb z}^p$ and $({\pmb W^p} {\pmb a^{p-1}} + {\pmb b}^{p} - \hat{\pmb z^p})^2 \approx({\pmb W^p} {\pmb a^{p-1}} + {\pmb b}^{p})^2$. The averaged energy in the $p$-th layer is
\begin{eqnarray}
\langle E_p \rangle &\sim& \frac{1}{n}\int \sum_{i=1}^n \sum_{k=1}^{l_p} \left(\sum_{j=1}^{l_{p-1}} W_{kj}^p a_{ij}^{p-1} + b_k^p - y_{ik}\right)^2 f({\pmb W}^L) f({\pmb b^L}) d{\pmb W^L} d{\pmb b^L} \nonumber \\
&\sim & \frac{1}{n} \sum_{i=1}^n \sum_1^{l_p} \left[\sum_{j=1}^{l_{p-1}}\sigma^2 (a_{ij}^{p-1})^2 + \sigma^2\right]
\end{eqnarray}

\paragraph{Tanh Activation Function.} We have
\begin{equation}
\langle E_p \rangle \sim \frac{1}{n} \sum_{i=1}^n l_p (l_{p-1} + 1) \sigma^2 \sim l_p (l_{p-1} + 1)\sigma^2,
\end{equation}
and the temperature of the $p$-th layer is
\begin{eqnarray}
T_p \sim \frac{\langle E_p\rangle}{l_p (l_{p-1}+1) \ln \sigma} \sim \left(\frac{\sigma^2}{\ln \sigma}\right)
\end{eqnarray}
for $p\geq 2$, while since $\langle E_1 \rangle \sim l_1 \overline{X^2}\sigma^2$ and 
\begin{equation}
T_1 \sim \frac{l_1 \overline{X^2}}{l_0 + 1} \left(\frac{\sigma^2}{\ln \sigma}\right),
\end{equation}
so the  asymptotic temperatures of layers are
\begin{equation}
T_1 \propto T_2 \sim T_3 \sim ... \sim T_L \label{equ_t_relation},
\end{equation}
which are independent from layers. 

\paragraph{Sigmoid Activation Function.} In this case, the energy of the $p$-th layer can be written as
\begin{equation}
\langle E_p \rangle \sim l_p (\xi_{p-1}l_{p-1} + 1)\sigma^2,
\end{equation}
where $\xi_{p-1}$ measures the fraction of $z^{p-1}$  tends towards one in the asymptotic case. And the temperature of the layer is
\begin{equation}
T_p  \sim \left(\frac{\xi_{p-1}l_{p-1}+1}{l_{p_1}+1}\right) \left(\frac{\sigma^2}{\ln \sigma}\right)\label{equ_nn_E0_6}.
\end{equation}
Although $\xi_p$ makes some fluctuation, since $T_p \propto \left(\frac{\sigma^2}{\ln \sigma}\right)$  still holds, we see the temperature relation Equation (\ref{equ_nn_E0_6}) also holds for the case of Sigmoid activation function. 

\paragraph{ReLU Activation Function.} Similar to Equation (\ref{equ_nn_E0_5}), we can derive the energy of the $p$-th layer as
\begin{equation}
\langle E_p \rangle \sim 2 \left[\prod_{k=1}^p \left(\frac{l_k}{2}\right)\right] \sigma^{2p} \overline{X^2},
\end{equation}
while the temperature in the layer is
\begin{equation}
T_p \sim \left[\prod_{k=1}^{p-1} \left(\frac{l_k}{2}\right)\right] \frac{\sigma^{2p} \overline{X^2}}{(l_{p-1}+1)\ln \sigma} \propto \left(\frac{\sigma^{2p}}{\ln \sigma}\right)\overline{X^2}\label{equ_nn_E0_3}.
\end{equation}
As a result, the temperatures in each layer have
\begin{equation}
T_1 \ll T_2 \ll T_3 \ll .. \ll T_L\label{equ_t_relation_2}.
\end{equation}

\subsection{Heat Engine Analogy and Work Efficiency}\label{section_nn_heat}

\begin{figure}[t]
    \centering
    \includegraphics[width=0.95\linewidth]{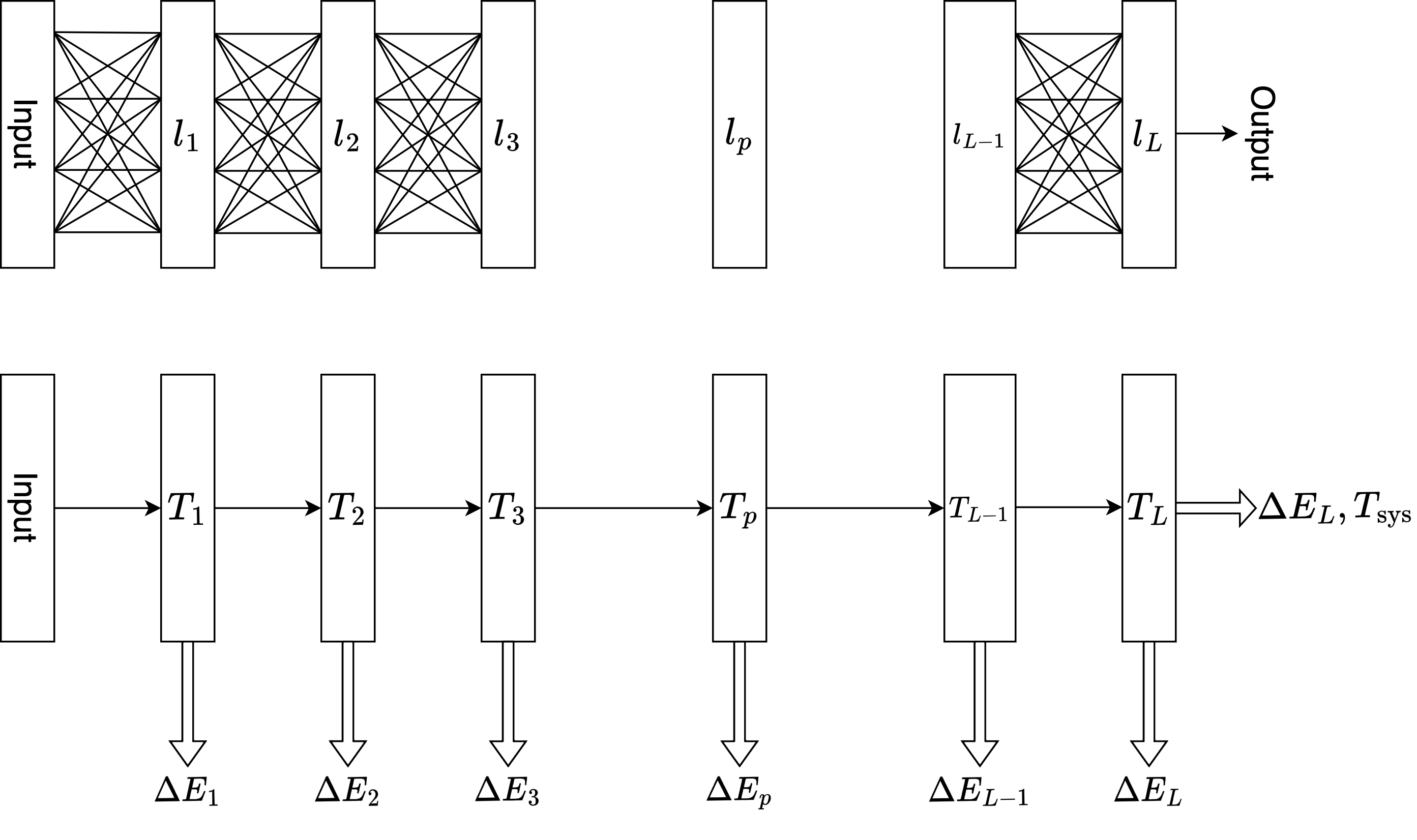}
    \caption{\small The structure of a typical artificial neural network. The upper figure shows the neural network's structure from a connectionism perspective, which includes input data and output result. In between, there are $L$ layers, with each layer's size being $l_1, l_2, l_3, ..., l_L$. The connection between two adjacent layers, i.e., the $p-1$ and the $p$-th layers, can be expressed using the weight matrix ${\pmb W^p} = \{W^p_{ij}\}$. On the other hand, the lower figure views the neural network from a physical perspective, considering it as a complex heat engine system, where each layer represents a part of this system. The $p$-th part of the system has a temperature $T_p$, and the released energy is $\Delta E_p$ . Only the energy released by the final part (the $L$-th layer) is useful externally and can be considered as the actual energy output of the system. From this we can calculate the work efficiency of this heat engine and compare the overall system temperature $T_{\rm syc}$ with the temperature of each part $\{T_p\}$.
}
    \label{fig_NN}
\end{figure}

The above results for neural networks can be interpreted from the perspective of thermodynamics and heat engine efficiency.  Figure \ref{fig_NN} shows a typical artificial neural network structure, we can view the system from a connectionist perspective (upper figure), or we can revisit this system from an energy perspective (lower figure).  The process of data moving from input to output through the system is also a process of the series of layers releasing energy: the $p$-th layer has a local temperature $T_p$ and releases energy $\Delta E_p$, while the eventually ``useful" energy released by the system (i.e. work done externally) is $\Delta E_L$, and the system temperature is $T_{\rm sys}$. 

We can define the work efficiency $\eta$ of the system as the ratio between work done externally and total energy released, i.e.
\begin{equation}
\eta = \frac{\Delta E_L}{\Delta E_{\rm tot}} = \frac{\Delta E_L}{\sum_{p=1}^L \Delta E_p}.
\end{equation}
The temperature of the system is
\begin{eqnarray}
T_{\rm sys} &=& \frac{\Delta E_L }{S_{\rm tot}} = \frac{\Delta E_L}{\Delta E_{\rm tot}} \frac{\Delta E_{\rm tot}}{S_{\rm tot}} \nonumber \\
&=& \eta \left(\frac{\sum_{p=1}^L T_pS_p}{\sum_{p=1}^L S_p}\right)\label{equ_t_sys_1}
\end{eqnarray}
If we assume that the entropy of each individual layer has the same order $S_1 \sim S_2 \sim ... 
\sim S_L$, thus $\eta = T_L/(\sum_p T_p)$, and the relation between temperatures in individual layers and overall system temperature is
\begin{equation}
T_{\rm sys} \sim \frac{T_L}{L} \sim \left(\frac{\eta}{L}\right) \sum_{p=1}^L T_p \sim \eta \overline{T},
\end{equation}
which shows that the overall system temperature is proportional to the average temperature of individual layers, with the work efficiency $\eta$ as the coefficient. 

\subsubsection{First Type of Heat Engine} \label{section_nn_physics_tanh}

For system using Tanh activation functions, we have
\begin{equation}\label{equ_nn_heat_uni}
	\left\{
	\begin{aligned}
		&\Delta E_1 = l_1 \overline{X^2}\sigma^2 , & S_1 = l_1 (l_0 + 1)\ln \sigma\\
		&\Delta E_2 = l_2 (l_1 + 1)\sigma^2 , & S_2 = l_2 (l_1 + 1)\ln \sigma \\
        &\Delta E_3 = l_3 (l_2 + 1)\sigma^2 , & S_3 = l_3 (l_2 + 1)\ln \sigma \\
        & ...... \\
        &\Delta E_L = l_L (l_{L-1} + 1)\sigma^2 , & S_L = l_L (l_{L-1} + 1)\ln \sigma
	\end{aligned}
	\right.
\end{equation}

The work efficiency of the system is
\begin{equation}
\eta = \frac{l_L (l_{L-1} + 1)}{l_1 \overline{X^2} + \sum_{p=1}^{L-1}l_{p+1} (l_p + 1)}.
\end{equation}
From Equation (\ref{equ_t_relation}) that $T_2 \sim T_3 \sim .. \sim T_L \sim T$ and Equation (\ref{equ_t_sys_1}), it is straightforward to obtain the temperature
\begin{eqnarray}
T_{\rm sys} &=& \frac{l_L (l_{L-1}+1)}{\sum_{p=0}^{L-1} l_{p+1} (l_p + 1)} T \nonumber \\
&=& \eta T \left[1 + \frac{S_1}{S_{\rm tot}} \left(\frac{T_1}{T} - 1 \right)\right]\label{equ_nn_heat_2}.
\end{eqnarray}
where $S_{\rm tot} = \sum_{p=1}^L S_p$. Note that Equation (\ref{equ_nn_heat_2}) also holds for system using Sigmoid activation function. For sufficient large total entropy or number of layers,  we can write $T_{\rm sys} \sim \eta T \sim \eta T_L$.

From a thermodynamic perspective, for systems where each layer has a comparable temperature (this cap is caused by the constraints of the activation functions), and the work efficiency of the system $\eta \ll 1$, and the system temperature $T_{\rm sys} \sim \eta T_L$, we refer to such neural network systems as the \textbf{First Type of Heat Engine}. 


\subsubsection{Second Type of Heat Engine}

On the other hand, for system using ReLU activation function, we have
\begin{equation}\label{equ_nn_heat_3}
	\left\{
	\begin{aligned}
		&\Delta E_1 = l_1 \sigma^2 \overline{X^2}, & S_1 = l_1 (l_0 + 1)\ln \sigma\\
		&\Delta E_2 = \frac{1}{2}l_2 l_1\sigma^4 \overline{X^2} , & S_2 = l_2 (l_1 + 1)\ln \sigma \\
        &\Delta E_3 = \frac{1}{2^2}l_3 l_2 l_1\sigma^6 \overline{X^2} , & S_3 = l_3 (l_2 + 1)\ln \sigma \\
        & ...... \\
        &\Delta E_L = \frac{1}{2^{L-1}}\left(\prod_{p=1}^L l_p\right)\sigma^{2L} \overline{X^2} , & S_L = l_L (l_{L-1} + 1)\ln \sigma
	\end{aligned}
	\right.
\end{equation}
Note that $\Delta E_1 \ll \Delta E_2 \ll ... \Delta E_L$ and  $T_1 \ll  T_2 \ll ...  T_L$. We have the work efficiency 
\begin{equation}
\eta = \frac{\frac{1}{2^{L-1}}\left(\prod_{p=1}^L l_p\right)\sigma^{2L}}{\sum_{k=1}^L \frac{1}{2^{k-1}}\left(\prod_{p=1}^k l_p\right)\sigma^{2k}} \approx 1.
\end{equation}
And 
\begin{eqnarray}
    T_{\rm sys} = \frac{T_L S_L}{\sum_{i=1}^L S_i} \sim \eta T_L\left(\frac{S_L}{S_{\rm tot}}\right).
\end{eqnarray}
For sufficient complex system with large number of layers, $T_{\rm sys} \ll \eta T_L$, which is different from the result in Section \ref{section_nn_physics_tanh} that $T_{\rm sys} \sim \eta T_L$.  Activation functions can be used to control the temperatures in the system. 

From a thermodynamic perspective, in principle there is no upper limit on the local temperature of the system. As the number of layers increases, the local temperatures gradually increase, the work efficiency $\eta$ of the system approaches 1 $\eta \approx 1$, and the system temperature $T_{\rm sys} \sim T_L S_L/S_{\rm tot}$. We refer to neural network systems with these properties as the as the \textbf{Second Type of Heat Engine}. 

\subsection{Other Parameter Initializations}\label{section_nn_others}

In the discussions from Section \ref {section_nn_normal} to Section \ref{section_nn_heat}, we assumed all the parameters initially follow normal distributions with the same standard deviation $\sigma$. However, in practice, there are different methods to initialize the parameters.

One common method is to use a different standard deviation for the initial normal distribution of parameters in each layer, for example, set $\sigma_p \propto \sqrt{1/l_{p-1}}$, or
\begin{equation}
  \sigma_p = \frac{\sigma_0}{\sqrt{l_{p-1}}}.
\end{equation}
So for the default case (Section \ref{section_nn_default}), the energy can be written as
\begin{eqnarray}
\langle E_0 \rangle \sim \frac{1}{n} \sum_i \frac{l_L (l_{L-1} + 1)}{l_{L-1}} \sigma_0^2 \sim \sigma_0^2 l_L \left(1 + \frac{1}{l_{L-1}}\right).
\end{eqnarray}
Note that for any layer the entropy $S_p  = l_p (l_{p-1} + 1) \ln \sigma_p = l_p (l_{p-1}+1) \ln \sigma_0 + \textrm{const.} \rightarrow l_p (l_{p-1} + 1)\ln \sigma_0$ for asymptotic case. Thus, the system temperature is
\begin{eqnarray}
T \sim \frac{l_L \sigma_0^2}{\sum_{p=1}^{L} l_p (l_p + 1) \ln \sigma_0}\left(1 + \frac{1}{l_{L-1}}\right) \sim \frac{l_L \sigma_0^2}{\sum_{p=1}^{L} l_p (l_p + 1) \ln \sigma_0}.
\end{eqnarray}
Compared to Equation (\ref{equ_nn_E0_2}), the relation $T \propto \frac{\sigma^2}{\ln \sigma}$ still holds, but the temperature is $1/(l_{L-1} + 1)$ of the original version.  For each layer, the local temperatures are
\begin{eqnarray}
T_p \sim \frac{\sigma_0^2}{l_{p-1} \ln \sigma_0}
\end{eqnarray}
for $p\geq 2$. We have $T_L \sim \frac{l_{L-2}}{l_{L-1}}T_{L-1} \sim \frac{l_{L-3}}{l_{L-1}}T_{L-2}\sim ... \sim \frac{l_1}{l_{L-1}}T_2$. The work efficiency of the system
\begin{equation}
\eta \sim \frac{l_L}{\frac{l_1}{l_0}\overline{X^2} + \sum_{p=2}^L l_p},
\end{equation}
and the system temperature is
\begin{eqnarray}
T_{\rm sys} = \eta T_L \frac{S_1 + \sum_{p=2}^{L}\frac{l_{L-1}S_p}{l_{p-1}}}{S_{\rm tot}} - \eta \Delta T \frac{S_1}{S_{\rm tot}},
\end{eqnarray}
where $\Delta T = T_L - T_1$. If $l_1 \sim l_2 \sim l_3 \sim ... \sim l_{L-1}$, 
\begin{equation}
T_{\rm sys} \sim \eta \left(T_L - \frac{S_1}{S_{\rm tot}}\Delta T \right) \sim \eta T_L,
\end{equation}
which is similar to Equation (\ref{equ_nn_heat_2}).

For the system with ReLU activation function (see Section \ref{section_nn_relu}), using $\sigma_p = \sigma_0/\sqrt{l_{p-1}}$, we obtain
\begin{eqnarray}
\langle E_0 \rangle \sim \frac{1}{2^{L-1}} \frac{l_L}{l_0} \sigma_0^{2L} \overline{X^2},
\end{eqnarray}
We still have
\begin{equation}
T \propto \left(\frac{\sigma_0^{2L}}{\ln \sigma_0}\right) \overline{X^2},
\end{equation}
but with a lower temperature compared with Equation (\ref{equ_nn_E0_3}). The local temperature in each layer is
\begin{equation}
    T_p \sim \frac{\overline{X^2}}{2^{p-1}l_0 (l_{p-1}+1)}\frac{\sigma_0^{2p}}{\ln \sigma_0},
\end{equation}
and $T_1 = \sigma_0^2 \overline{X^2}/[l_0 (l_0 + 1)\ln \sigma]$. Such, we have $T_1 \ll T_2 \ll T_2 \ll ... \ll T_L$, the work efficiency $\eta \approx 1$, and 
\begin{eqnarray}
    T_{\rm sys} \sim \eta T_L \left(\frac{S_L}{S_{\rm tot}}\right),
\end{eqnarray}
The key results for Type II heat engine still holds. 

As a conclusion, if the initial normal distribution of the parameters is different, the relation between the system temperature and the local temperatures in the first type of heat engine varies, and the system's efficiency also differs slightly. For the second type of heat engine, the expressions for system efficiency and temperature remain the same.

If the initial setting of the system parameters is uniform distribution, the aforementioned conclusions still hold. We discuss the case where the parameters are initially set to a uniform distribution in the Appendix \ref{appendix_nn}.

\section{Conclusions}

We develop a thermodynamic framework for machine learning (ML) systems. Similar to physical thermodynamic systems, which are characterized by energy and entropy, ML systems also possess these characteristics. This comparison inspired us to integrate the concept of temperature into ML systems, grounded in the fundamental principles of thermodynamics, and to establish a basic thermodynamic framework for machine learning systems with non-Boltzmann distributions.

We first introduce the concept of states into ML systems. We propose three fundamental elements of a basic ML system: the model (with parameter set $\pmb \mu$), data ($\mathcal {D} = \mathcal X \times Y$), and energy ($E$) (see Figure \ref{fig_ml_system}). ML systems have two types of states. For a given dataset, all possible ${\pmb \mu}$ in the parameter space forms a state, known as the \textbf{Type I State}.  This state represents an ML system that has not yet been trained, where each $\{E_i, \pmb \mu_i\}$ can be considered as a particle. After model training, all the particles converge to $\{\hat{E}, \hat{\pmb \mu}\}$, and the transition from all $\{E_i, \pmb \mu_i\}$ to converged $\{\hat{E}, \hat{\pmb \mu}\}$ can be considered an isothermal phase transition, allowing us to calculate the corresponding temperature of the system. On the other hand, unlike the Type I state, the \textbf{Type II state} of an ML system is defined by the dataset after specifying a set of parameters $\hat{\pmb  \mu}$. The transformation of the Type II state corresponds to shifts in ML data and the continuous refresh of the model.

After providing a comprehensive discussion and unified scenario on global temperature of ML systems, we primarily discuss the temperature of Type I state in phase transitions. We consider that the initial potential energy of an ML system is described by the model loss functions, and the energy adheres to the principle of minimum potential energy. Regarding the probability distribution $f({\pmb \mu})$ within the parameter space, we argue that the temperature of the Type I state is (Equation \ref{equ_T_diff})

\begin{equation}
T = \frac{ E_f - \int E({\pmb \mu})f({\pmb \mu})d{\pmb \mu}}{\int f({\pmb \mu}) \log [f({\pmb \mu})]d{\pmb \mu}} \nonumber.
\end{equation}
It is important to clarify that the system energy we propose is fundamentally different from the energy in EBMs. Within the EBM framework, both the loss function and the energy function coexist. The energy function is minimized by the inference process, while the loss functional is minimized by the learning process.  In our theoretical framework, we do not distinguish between the loss function and the energy function.

Next, we derive the temperatures of a various ML toy systems, which have models with initial parameters following either normal or uniform distributions, while their energy forms include linear regression with Mean Squared Error (MSE) or Mean Absolute Error (MAE) and regularization, as well as logistic regression with cross entropy. Certain system temperatures under certain inertial parameter distributions and energy forms have analytical solutions. For example, for an initial normal distribution of parameters and linear regression, the system temperature is (Equation [\ref{equ_gauss_mse_highD}])
\begin{eqnarray}
T = \frac{\sum_{j=1}^{K} \sigma_j^2 \overline{X_j^2} + \overline{Y^2} -  \overline{{\rm tr}\{C_x - C_{xy}C_y^{-1} C_{yx}\}}}{\sum_{j=1}^K \ln \sigma_j + \frac{K}{2}[1+ \ln(2 \pi)]}\nonumber,
\end{eqnarray}
where $K$ is the dimension of the parameter space, and $X_{i}$ is the $i$-th component of the data point $\pmb X$. In more general cases, we consider the asymptotic solutions for the system temperatures, in particular for normal distributions $\{\sigma_i\}\rightarrow \sigma \rightarrow \infty$, or uniform distributions length $\{l_i\}\rightarrow l \rightarrow \infty$, Table \ref{tab_solutions} gives the asymptotic solutions. 

\begin{table}[ht]
\centering
\footnotesize
\begin{tabular}{lcl}
\hline
\textbf{Energy Form} & \textbf{Initial ${\pmb \mu}$ Distribution}& \textbf{Solution} \\
\hline
\hline
LineR + MSE& normal & \(\displaystyle
T \sim  \overline{X^2} \left(\frac{\sigma^2}{\ln \sigma}\right) \) \\
\hline
LineR + MSE& uniform & \(\displaystyle
T \sim \overline{X^2} \left(\frac{l^2}{12\ln  l}\right) \) \\
\hline
LineR + MSE + reg& normal& $\begin{aligned}[t]
       & T_{L_1} \sim \overline{X^2}\left(\frac{\sigma^2}{\ln \sigma}\right) \\
       & T_{L_2} \sim (\overline{X^2} + \lambda) \left(\frac{\sigma^2}{\ln \sigma}\right)
      \end{aligned}$ \\
\hline
LineR + MSE + reg& uniform& $\begin{aligned}[t]
       & T_{L_1} \sim \overline{X^2} \left(\frac{l^2}{12\ln l}\right)  \\
       & T_{L_2} \sim (\overline{X^2} + \lambda) \left(\frac{l^2}{12\ln l}\right)
      \end{aligned}$ \\
\hline
LineR + MAE& normal& \(\displaystyle T  \sim \sqrt{\frac{2}{\pi}}  \left(\frac{\sigma}{\ln \sigma}\right) \overline{|X|}  \) \\
\hline
LineR + MAE& uniform& \(\displaystyle T \sim \frac{1}{4} \left(\frac{l}{\ln l }\right) |\overline{X}|  \) \\
\hline
LogR + CE& normal& \(\displaystyle
   T \propto \frac{\textrm{e}^{\sigma^2/2}}{\ln \sigma} \)  \\
\hline
LogR + CE& uniform& \(\displaystyle T \propto \frac{e^{l/2}}{l \ln l} \)  \\
\hline
NN + Tanh/Sigmoid& normal & \(\displaystyle T \sim \frac{l_L (l_{L-1}+ 1)}{\sum_{p=1}^{l_L} l_p (l_{p-1} + 1)}  \left(\frac{\sigma^2}{\ln \sigma}\right) \)  \\
\hline
NN + ReLU & normal & \(\displaystyle T \propto \left(\frac{\sigma^{2L}}{\ln \sigma}\right) \overline{X^2} \)  \\
\hline
\end{tabular}
\caption{\small The asymptotic temperatures of various ML systems. The energy forms of the systems include linear regression (LineR) with MSE and MAE and regularization (reg), logistic regression (logR) with Cross Entropy (CE), and neural network (NN) with Tanh, Sigmoid and ReLU activation functions. The initial parameter (${\pmb \mu}$) distributions for the systems are either normal or uniform.}\label{tab_solutions}
\end{table}

We also give a physical explanation of the ML temperature. If two systems, $A$ and $B$, have the same data distribution $\overline{X_A^2} = \overline{X_B^2}$, they also share the same temperature. From a physical perspective, the two systems are in equilibrium. The process of mixing non-equilibrium systems to achieve  equilibrium essentially involves the transfer of energy from the higher-temperature system to the lower-temperature system. The scenarios discussed above regarding temperature changes in equilibrium, non-equilibrium, and model retraining systems bear a strong resemblance to traditional thermodynamic systems. This forms the thermodynamic landscape of temperature in ML systems.

Next we discuss ML systems with neural network models. For a neural network with $L$ layers, we propose that each layers of a neural network has its own local temperature $T_i$  ($i=1,2,3,...,L$), also the system has its global temperature $T_{\rm sys}$. We found that the relationship between the local temperatures of a neural network depends on the model activation function. For example, with Tanh and Sigmoid activation functions, we have $T_1 \propto T_2 \sim T_3 \sim ... \sim T_L$ (Equation [\ref{equ_t_relation}]), while for ReLU activation function, we have $T_1 \ll T_2 \ll T_3 \ll .. \ll T_L$ (Equation [\ref{equ_t_relation_2}]).  These results for neural networks can be interpreted from the perspective of thermodynamics and \textbf{heat engine efficiency}. Viewing a neural network system as a \textbf{heat engine} (see Figure \ref{fig_NN}), where the energy output of each layer is $\Delta E_i$, we can define the efficiency of this heat engine system as
\begin{equation}
\eta = \frac{\Delta E_L}{\Delta E_{\rm tot}} = \frac{\Delta E_L}{\sum_{p=1}^L \Delta E_p}.
\end{equation}
We then classify neural networks based on their work efficiency, identifying them as one of two types of heat engines. For systems using Tanh activation functions, we observe that $T_1 \propto T_2 \sim T_3 \sim ... \sim T_L \sim T$,  $\eta \ll 1$ and the system tempeature $T_{\rm sys} \sim \eta T_{L} $. We define such low-efficient neural network systems as the \textbf{First Type of Heat Engine}. Conversely, for systems employing the ReLU activation function, we find that $\eta \approx 1$, and the system temperature $T_{\rm sys} \ll \eta T_{L}$. We refer to neural network systems with these properties as the as the \textbf{Second Type of Heat Engine}. 

\newpage

\appendix

\section{Appendix I: The Unified Scenario for Type I and Type II States}\label{appendix_I}

\begin{figure}[p]
    \centering
    \includegraphics[width=0.85\linewidth]{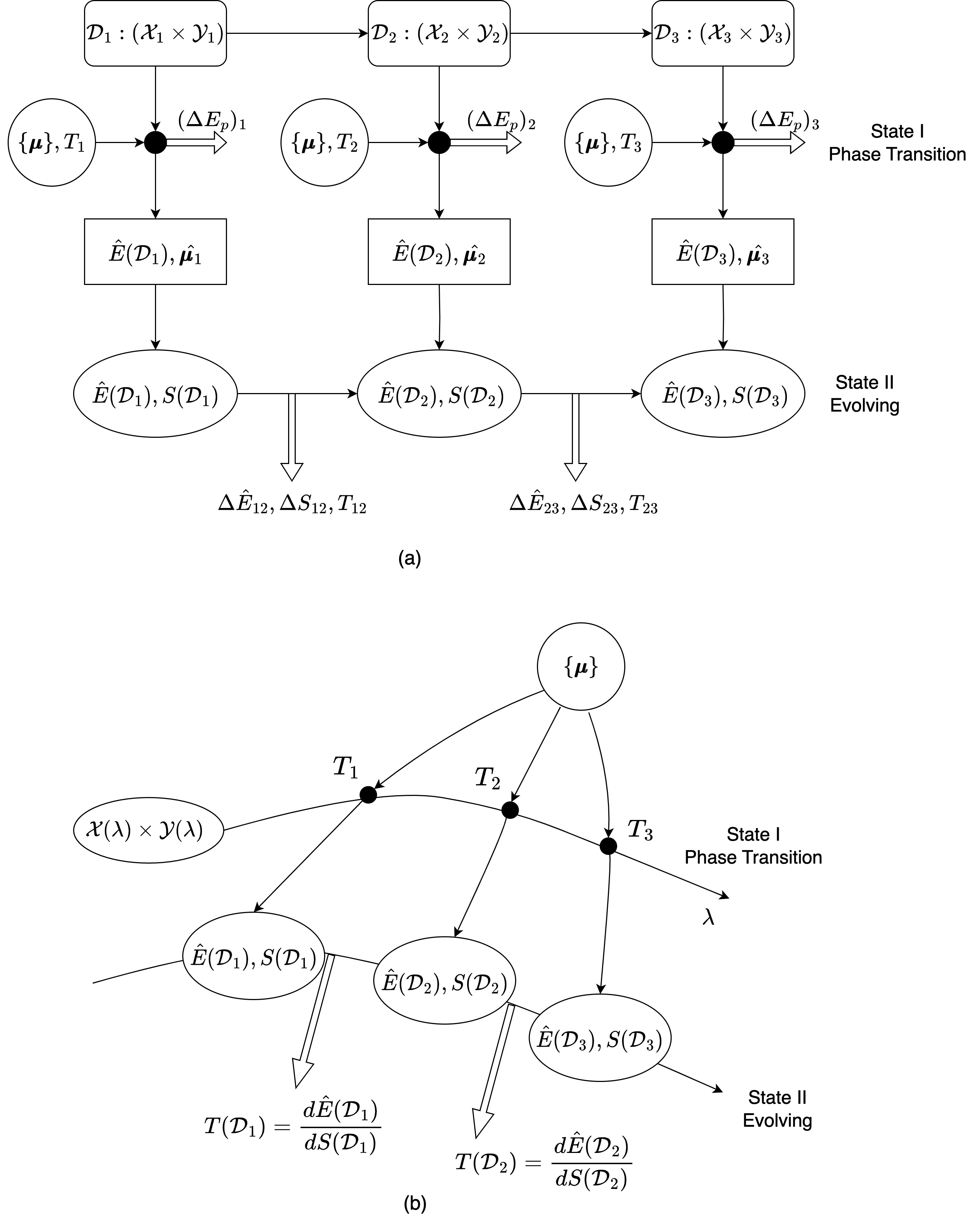}
    \caption{The entire process of training and retraining a ML system from a thermodynamic perspective, which shows a unified scenario to combine Type I and Type II states from Section \ref{section_state} for discrete case (upper figure (a)) and continuous case (lower figure (b)). The black dots in upper figures represent Type I state phase transition. This figure develops the idea in Figure \ref{fig_intro}. For detailed discussion of the figure see Appendix \ref{appendix_I}.}
    \label{fig_unified_states}
     \vspace{-5pt}
\end{figure}

For Type II State from Section \ref{section_state}, we need to calculate the system entropy through the joint distribution $P(\mathcal{X}, \mathcal{Y})$ in equation (\ref{equ_stateII_entropy}), which can be normalized by
\begin{equation}
P(x_i, y_i) = \frac{P(x_i, y_i)}{\sum_{{(x_i, y_i)} \in \mathcal{D}} P(x_i, y_i)},
\end{equation}
However, the probability distribution of the input and output data domain for most ML systems is unknown. We can only estimate the distribution via some models, i.e., using generative models to estimate $P(\mathcal{X}, \mathcal{Y})$, or discriminative models to estimate the conditional probability $P(\mathcal{Y}|\mathcal{X})$, and we need to know or assume the probability distribution of $P(\mathcal{X})$, so we can estimate $P(\mathcal{X}, \mathcal{Y})$ by $P(\mathcal{Y}|\mathcal{X})P(\mathcal{X})$.

Figure \ref{fig_unified_states} shows the entire scenario of combining Type I and Type II states, where the upper figure (a) illustrates the discrete case. Suppose the initial system (training) dataset is $\mathcal{D}_1: \mathcal{X}_1 \times \mathcal{Y}_1$, and the model parameter space is $\{\pmb \mu\}$. The fixed dataset $\mathcal{D}_1$ and the parameter space $\{\pmb \mu\}$ with a certain distribution gives a Type I state of the system. The ML training process is equivalent to the phase transition process into a new state with energy $\hat{E}(\mathcal{D}_1)$ and with an optimized (fixed) parameter set  $\hat{\pmb \mu_1}$, with energy $\Delta (E_p)_1$ released from phase transition. We can use Equation (\ref{equ_state_I_T}) to calculate the corresponding temperature $T_1$ through the Type I phase transition. Next, through the model  $\hat{f}$ corresponding to $\hat{\pmb \mu_1}$, we can compute the conditional probability $P(\mathcal{Y}_1 | \mathcal{X}_1)$ on dataset $\mathcal{D}_1$, and obtain an approximation of $P(\mathcal{X}_1, \mathcal{Y}_1) = P(\mathcal{Y}_1 | \mathcal{X}_1) P(\mathcal{X}_1)$ by assuming the distribution of $P(\mathcal{X}_1)$. Thus, we can then use Equation (\ref{equ_stateII_entropy}) to calculate the corresponding entropy $S(\mathcal{D}_1)$, which is the data entropy of the Type II state.  

As the system data varies from $\mathcal{D}_1$ to $\mathcal{D}_2 = \mathcal{X}_2 \times \mathcal{Y}_2$, the Type I state changes to temperature $T_2$, and undergoes a phase transition into a state with energy $\hat{E}(\mathcal{D}_2)$ with an updated optimized parameter set $\hat{\pmb \mu_2}$, and the data entropy of the system changes to $S(\mathcal{D}_2)$. Using Equation (\ref{equ_T12}), we can calculate the corresponding temperature of Type II state. As shown in Figure \ref{fig_unified_states}, this calculation can be continued as the training dataset sequence $\{\mathcal{D}_k\}$ evolves. 

The lower panel of Figure \ref{fig_unified_states} shows the continuous case of system evolution, where the system data $(\mathcal{X}(\lambda), \mathcal{Y}(\lambda))$ changes continuously. We can still start from the model parameter initialization to form a Type I state  $\mathcal{D}_k = \mathcal{X}(\lambda_k) \times \mathcal{Y}(\lambda_k)$ at a certain $\lambda_k$ point, and this state undergoes a phase transition through ML training, with its temperature and energy being $T_k$ and $\hat{E}(\mathcal{D}_k)$ respectively. We can then calculate the data entropy of the system after the phase transition from the perspective of Type II state. The temperature of Type II state can be calculated as
\begin{equation}
T(\mathcal{D}_k) = \frac{d \hat{E}(\mathcal{D}_k)}{d S(\mathcal{D}_k)}.
\end{equation}  

In short, Figure \ref{fig_unified_states} combines Type I and Type II states, describing the entire process of training and retraining a ML system from a thermodynamic perspective.

\section{Appendix II: Entropy Change and Dimension Collapse}\label{appendix_entropy_change}

Let us begin with a simple example. We consider a two-dimensional (2D) uniform distribution in a rectangle region that
\begin{equation}
f(x, y) = \frac{1}{ab}
\end{equation}
for $x_0\leq x \leq x_0 + a$ and $y_0\leq y \leq y_0 + b$, otherwise $f(x,y) = 0$. The differential entropy of the uniform distribution is given by
\begin{equation}
S_2(x,y) = \log(ab)\label{equ_2D_uniform}.
\end{equation}
Next, we squeeze the $y$-direction of the 2D region such that $b$ approach a small number $\delta$. Then the differential entropy Equation (\ref{equ_2D_uniform}) becomes 
\begin{equation}
S_\delta(x,y) = \log a + \log \delta \label{equ_2D_uniform_2}.
\end{equation}
As $\delta \rightarrow 0$, the 2D probability distribution actually becomes a 1D (line) distribution.  From the 1D perspective, the entropy of the new distribution is
\begin{equation}
S_1(x) = \log a \label{equ_1D_uniform}.
\end{equation}
Clearly, the term $\log \delta$ in Equation (\ref{equ_2D_uniform_2}) represents the entropy change resulting from dimensional collapse from two dimensions to one dimension, while $S_2(x,y) - S_1(x) = \log b$ is the actually entropy from 1D. 

Using the same logic, let us re-examine the connection between differential entropy and discrete entropy in Section \ref{section_sys_entropy}. For a probability distribution function $f({\pmb \mu})$, we assume the parameter space of ${\pmb \mu}$ is divided into a series of small grids $\{\Delta_i\}$, and the probability distribution collapses into a Dirac delta function $\delta_D ({\pmb \mu - \pmb \mu_i})$ within each grid $\Delta_i$ and $f({\pmb \mu_i}) = p_i$, we see the entropy change as
\begin{eqnarray}
S_{\rm diff} &=& -\int f({\pmb \mu}) \log[f({\pmb \mu})] d{\pmb \mu} \nonumber \\
&\rightarrow& -\sum_i \int_{\Delta_i} f({\pmb \mu}) \delta_D ({\pmb \mu - \pmb \mu_i}) \log [f({\pmb \mu}) \delta_D ({\pmb \mu - \pmb \mu_i})] d{\pmb \mu} \nonumber \\
& = & -\sum_i f({\pmb \mu_i}) \log[f({\pmb \mu_i}) \delta(0)] = -\sum_i p_i \log[p_i \delta_D^{\mathcal{N}}(0)] \nonumber \\
& = & S_{\rm discrete} + \mathcal{N}\log \delta ,
\end{eqnarray}
where we denote $\delta({\pmb \mu_i} - {\pmb \mu_i}) = [\delta_D(0)]^{\mathcal{N}} = (1/\delta)^{\mathcal{N}}$ with $\mathcal{D}$ being the dimension of $\pmb \mu$ and $\delta \rightarrow 0$ as shown in Equation (\ref{equ_2D_uniform_2}). The change in entropy, in addition to the change from differential entropy $S_{\rm diff}$ to discrete entropy $S_{\rm dicrete}$, also includes the entropy change due to dimensionality reduction $\mathcal{N}\log \delta$. 

In general, for entropy change from dimension $\mathcal{D}_1$ to $\mathcal{D}_2$ (where $\mathcal{D}_1 > \mathcal{D}_2$), the overall entropy change is written as 
\begin{equation}
\Delta S = S_{\mathcal{D}_2} - S_{\mathcal{D}_1} + (\mathcal{D}_1 - \mathcal{D}_2) \log \delta \label{equ_entropy_change},
\end{equation}
where $S_{\mathcal{D}_1}$ and $S_{\mathcal{D}_2}$ are entropy in the corresponding dimensional spaces. $\Delta S_1 = S_{\mathcal{D}_2} - S_{\mathcal{D}_1}$ shows the ``real" finite entropy change,  and $(\mathcal{D}_1 - \mathcal{D}_2) \log \delta$ is caused by dimension collapse. For a probability distribution in the space $\pmb \mu$ collapses to a single point $\mu_0$, we have ${\mathcal D_1} = \mathcal N$, $\mathcal D_2 =0$,  $S_{\mathcal D_1} = S_{\rm diff}$, $S_{\mathcal D_2} = S_{\rm discrete}$, and in the zero-dimension space with a single point, there is zero entropy that $S_{\rm discrete} = 0$. So from the general Equation (\ref{equ_entropy_change}) we can use
\begin{equation}
\Delta S_1 = 0 - S_{\rm diff} = -S_{\rm diff}
\end{equation}
to represent the change in entropy, while $\mathcal{N} \log \delta $ corresponds to the entropy change due to dimensionality reduction.

For a ML system with Type I state phase transition, given that the energy change $\Delta E$ is a finite quantity, we also use a finite quantity $S_{\rm diff}$ to express the change in entropy. Thus, the temperature of Type I state phase transition $T = (E_f - E_0)/(-S_{\rm diff}) = (E_0 - E_f)/S_{\rm diff}$. This is exactly Equation (\ref{equ_T_1}) we showed in Section \ref{section_state}, also Equation (\ref{equ_T_diff}) in Section \ref{section_sys_entropy}. 

Above we discussed the change in system entropy from the perspective of dimensional collapse. Next, we will look into the issue of infinite entropy from the perspective of a physical thermodynamic system. Suppose a system is made up by two subsystems. The energy change is
\begin{equation}
\Delta E = \Delta Q_1 + \Delta Q_2 = T_1 \Delta S_1 + T_2 \Delta S_2.
\end{equation}
The total entropy change in the system is $S_{\rm diff} - \log \delta$ (see Equqation [\ref{equ_S_discrete}]). We allocate $\Delta S_1 = -\log \delta$, $\Delta S_2 = S_{\rm diff}$, and $\Delta Q_1 = \delta$, $\Delta Q_2 = \Delta E - \delta$. So the temperature of the subsystem I is 
\begin{eqnarray}
T_1 &=& \frac{\delta}{ - \log \delta} = \frac{\delta }{\sum_{k =1}^{\infty}\frac{(1-\delta)^k}{k}} \approx \frac{\delta}{\sum_{k=1}^{\infty} \frac{1}{k}} \nonumber \\
&\sim& \frac{\delta}{{\rm lim}_{{k \rightarrow \infty}}\ln k + 0.57721} \rightarrow 0,
\end{eqnarray}
On the other hand, the temperature of the subsystem II is
\begin{equation}
T_2 = (\Delta E - \delta)/S_{\rm diff} \approx \Delta E/S_{\rm diff},
\end{equation}
which is the temperature we calculate in this paper. 

Note that if $S_{\rm diff} < 0$, $T_2$ can be negative. Negative thermodynamic temperature has been widely discussed in statistical and condense physics so this is not an unfamiliar concept \cite{Onsager49, Ramsey56}.

\section{Appendix III: Non-Boltzmann and Generalized Boltzmann Distributions}\label{appendix_non_boltzmann}

We can derive the Boltzmann distribution of a gaseous system from first principles \cite{Landau_stats}. If the gaseous system is in a constant gravitational potential with acceleration $g$, for a gas particle with velocity $v$ and at a height $h$, we have\begin{equation}
\varepsilon_{v,h} = \frac{1}{2} mv^2 + mgh \label{appendix_energy_1}.
\end{equation}
The configuration is subject to the constraints
\begin{eqnarray}
&&\iint n_{v,h}dvdh = N \\
&&\iint n_{v,h} \varepsilon_{v, h} dv dh = E,
\end{eqnarray}
where $N$ and $E$ are total number of particles and energy respectively. The number of ways of making a given configuration is
\begin{equation}
\Omega(\{n_{v_i,h_j}\}) = \frac{N!}{\prod_{i,j} n_{v_i,h_j}!}.
\end{equation}
We want to maximize $\Omega$, therefore we have
\begin{equation}
\frac{\partial}{\partial n_{v,h}} (\ln \Omega - \beta E - \alpha N) = 0
\end{equation}
\begin{equation}
\frac{n_{v,h}}{N} = \textrm{e}^{-\alpha - \beta \varepsilon_{v, h}}\label{B_dist_1},
\end{equation}
Using equation (\ref{appendix_energy_1}), equation ({\ref{B_dist_1}}) can be written as
\begin{equation}
n_{v,h}  \propto \exp\left(-\alpha - \frac{mv^2 + 2mgh}{2k_B T }\right),
\end{equation}
where we can prove $\beta=1/k_B T$ using thermodynamic relations.

The Boltzmann distribution has already been extended to the cases of special and general relativity \cite{Juttner1911,Walker1936, Kremer12, Zaninetti2020}. However, how to establish the theories of thermodynamics and statistical mechanics within the framework of relativity remains an open question \cite{Rovelli13, Lopez_Monsalvo2011, Mendes2021}. In this appendix, we continue to discuss within the framework of classical mechanics. We have given the above calculations in the rest frame stationary relative to the ground. Next, let us consider an observer moving relative to the ground with velocity ${\pmb V}$. For this observer, a gas particle with velocity $\tilde{\pmb v}$ would appear to have a velocity $\tilde{\pmb v} - {\pmb V}$ in the observer's reference frame. We can establish the velocity transformation between the two frames of reference:
\begin{equation}
({\rm rest\, frame})\;\tilde{\pmb v} \;{\longleftrightarrow}\; {\pmb v} = \tilde{\pmb v} - {\pmb V}\;({\rm observer's\,frame}),
\end{equation}
Then, from the observer's frame of reference,
\begin{eqnarray}
n_{v,h} &\propto& n_{\tilde{v},h} \propto \exp\left(-\alpha - \frac{m\tilde{v}^2 + 2mgh}{2k_B T }\right) \nonumber \\
&\propto& \exp\left[-\frac{m({\pmb v} + {\pmb V})^2 + 2mgh}{2k_B T}\right] \nonumber \\
&\propto& \exp\left(\frac{mV^2}{2k_B T}\right) \exp\left( - \frac{mv^2 + 2mgh}{2k_B T }\right) 
\left\langle \exp(- \frac{m}{k_B T} {\pmb v}\cdot{\pmb V})\right\rangle \nonumber \\
&\propto&  \exp\left(\frac{mV^2}{2k_B T}\right) \exp\left( - \frac{\varepsilon_{v, h}}{2k_B T }\right)\label{B_dist_3}. 
\end{eqnarray}
We can also start from the first principles. From the observer's frame, the energy in the gaseous system gives
\begin{eqnarray}
\iint n_{v,h} \varepsilon_{v, h} dvdh &=& \iint \left[\frac{1}{2} m (\tilde{\pmb v} - {\pmb V})^2  + mgh\right] n_{v,h} dvdh \nonumber \\
& = & \iint \left(\frac{1}{2}m \tilde{v}^2 + mgh\right) n_{\tilde{v},h}d\tilde{v}dh -  \iint m {\bf v} \cdot {\bf V}n_{\tilde{v},h} d\tilde{v}dh d\omega \nonumber \\
&+& \frac{1}{2} mV^2  \iint n_{v,h} dvdh \nonumber \\
& = & E + \frac{1}{2} mV^2 \iint n_{v,h}dvdh,
\end{eqnarray}
Therefore we have two constraints in the observer's frame that
\begin{eqnarray}
&&\iint n_{v,h}dvdh = N \\
&&\iint n_{v,h} \varepsilon_{v, h} dv dh - \frac{1}{2} mV^2 \iint n_{v,h}dvdh = E,
\end{eqnarray}
Thus we can obtain
\begin{equation}
\ln\left(\frac{n_{v,h}}{N}\right) - \alpha - \beta\left[\varepsilon_{v, h}  - \frac{1}{2}mV^2\right] = 0,
\end{equation}
or
\begin{equation}
n_{v,h} \propto \exp\left(\frac{1}{2}\beta m V^2\right) {\rm e}^{-\beta \varepsilon_{v, h}},
\end{equation}
which is the same as equation (\ref{B_dist_3}). Clearly, $n_{v,h}$ differs from the traditional Boltzmann distribution by a factor of $f(V) = \frac{1}{2} mV^2$. If $V$ varies, then $n_{v,h}$ also redistributes with $V$. this can be referred to as a non-Boltzmann distribution or a generalized Boltzmann distribution. 

Some research divides the energy of a gaseous system into internal energy and bulk energy. However, from the perspective of relativity, there is no special reference frame among various reference frames. Therefore, setting a specific center-of-mass reference from is also unreasonable, especially in many cases, we cannot accurately determine the velocity of the center of mass, nor split energy of each gas particles to internal and bulk energy. 

In the above discussion, we assumed that the gravitational potential energy is a constant. We can consider more general cases. We split the energy to internal and others for the $i$-th state of a system:
\begin{equation}
\varepsilon_ i = \hat{\varepsilon_i} + \mathcal{G}_i,
\end{equation}
where $\mathcal{G}_i$ is contributed by gravitational potential and other input/output energies. And we assume the internal energy is conserved, so we have
\begin{eqnarray}
\int n_i \varepsilon_ i  d\Omega &=& \int n_i \hat{\varepsilon_ i} d\Omega + \int n_i \mathcal{G}_i d\Omega  \nonumber \\
& = & E + \int n_i \mathcal{G}_i d\Omega,
\end{eqnarray}
so the constraints are given by
\begin{eqnarray}
&& N = \int n_i d\Omega \\
&& E = \int n_i \varepsilon_i d\Omega - \int n_i \mathcal{G}_i d\Omega.
\end{eqnarray}
The distribution of the system is 
\begin{equation}
\ln \left(\frac{n_i}{N}\right) - \alpha - \beta \varepsilon_i  + \beta \mathcal{G}_i = 0,
\end{equation}
i.e., 
\begin{eqnarray}
n_i &\propto& \exp\left(-\alpha - \beta \varepsilon_i + \beta \mathcal{G}_i\right)  \nonumber \\
& \propto & f_i(\mathcal{G}) {\rm e}^{-\alpha - \beta \varepsilon_i }\label{partition_metric}.
\end{eqnarray}
Here, the factor $f_i(\mathcal{G}) = {\rm e}^{\beta G_i}$ can be called the \textbf{metric of the partition function}. Therefore, Boltzmann distribution is just s special cases for a system without varying gravitational background. And Equation (\ref{partition_metric}) can be called the generalized Boltzmann distribution.

\section{Appendix IV: Temperature and Thermodynamics Laws with Lorentz Transformation}\label{appendix_lorentz}

In the framework of relativity, how does temperature transform between difference reference frames? This remains an open question and has been debated for a long time. Assume a body is measured with a temperature $T$ in the rest frame, while its temperature is $T'$ in another frame where the body has a velocity $v$. How do we establish the Lorentz-like transformation from $T$ to $T'$? There have been three prevailing viewpoints:

\begin{itemize}

\item The observed moving body has the temperature $T' = T/\gamma$ \cite{Planck1907, Einstein1907}, where $\gamma=(1-v^2/c^2)^{-1/2}$ is the Lorentz factor. 

\item The moving body has the temperature $T' = T\gamma$ \cite{Otto1963, Arzelies1965}.

\item The body temperature is an invariant that $T' = T$ \cite{Landsberg1966}. 

\end{itemize}

Discussing whether the temperature $T$ of a body is a Lorentz invariant is beyond the scope of this paper. However, many studies have considered the thermodynamic relation
\begin{equation}
T dS = dU + PdV
\end{equation}
to be Lorentz covariant, which means the fundamental thermodynamic relation holds in all reference frames. Based on this, we can still use
\begin{equation}
T = \left(\frac{\partial  U}{\partial S}\right)_{V}
\end{equation}
to calculate temperature in any reference frames.

\section{Appendix V: Neural Network with Initially Uniform Distributed Parameters} \label{appendix_nn}

All discussions in Section \ref{section_nn}  are based on the condition that the initial parameters follow normal distributions. In this appendix, we discuss the scenario where the parameters of the neural network follow uniform distributions. Assuming all parameters are randomly distributed between $[-l/2, l/2]$, the asymptotic energy of the system with $l \rightarrow \infty$ and Tanh activation function is
\begin{eqnarray}
\langle E_0 \rangle &=& \frac{1}{n}\int \sum_{i=1}^n \sum_{k=1}^{l_L} \left(\sum_{j=1}^{l_{L-1}} W_{kj}^L a_{ij}^{L-1} + b_k^L - y_{ik}\right)^2 f({\pmb W}^L) d{\pmb W^L} db^L \nonumber \\
&=& \frac{1}{n}\sum_i \sum_k \left[\sum_j \frac{l^2}{12} (a_{ij}^{L-1})^2 + \frac{l^2}{12} + (y_{ik})^2\right]\nonumber \\
&\sim& \frac{l^2}{12} l_L (l_{L-1} + 1).
\end{eqnarray}
From this, the temperature of the system is 
\begin{equation}
T_{\rm sys} \sim \frac{l_{L-1} + 1}{\sum_{p=1}^L l_p (l_p+1)} \left(\frac{l^2}{12 \ln l}\right)
\end{equation}
For the energy and temperature of each layer, we can calculate as follows
\begin{equation}
	\left\{
	\begin{aligned}
		&\Delta E_1 = l_1 \frac{l^2}{12} \overline{X^2}, & S_1 = l_1 (l_0 + 1)\ln l\\
		&\Delta E_2 = l_2 (l_1 + 1)\frac{l^2}{12} , & S_2 = l_2 (l_1 + 1)\ln l \\
        &\Delta E_3 = l_3 (l_2 + 1)\frac{l^2}{12} , & S_3 = l_3 (l_2 + 1)\ln l \\
        & ...... \\
        &\Delta E_L = l_L (l_{L-1} + 1)\frac{l^2}{12} , & S_L = l_L (l_{L-1} + 1)\ln l
	\end{aligned}
	\right.
\end{equation}
And we obtain $T_2 \sim T_3 \sim T_4 \sim ... \sim T_L \sim l^2/\ln l $, which belongs to the  first type of heat engine. Clearly, by simply substituting 
\begin{equation}
\frac{l^2}{12} \rightarrow \sigma^2,
\end{equation}
the energy and temperatures of the system in mathematical form is identical to that in Section \ref{section_nn}. For the case where the activation function is ReLU, the system remains the second type of heat engine, corresponding to the work efficiency $\eta \approx 1$ and $T_{\rm sys} \approx T_L (S_L/S_{\rm tot})$. Other details will not be elaborated further.
 


\bibliographystyle{unsrt}
\bibliography{references}

\end{document}